\documentclass[lettersize,journal]{cls/IEEEtran}

\usepackage{amsmath,amsfonts}
\usepackage{algorithmic}
\usepackage{algorithm}
\usepackage{array}
\usepackage[caption=false,font=footnotesize,labelfont=sf,textfont=sf]{subfig}
\usepackage{textcomp}
\usepackage{stfloats}
\usepackage{verbatim}
\usepackage{graphicx}
\usepackage{float}
\usepackage{placeins}
\usepackage[acronym]{glossaries}
\usepackage[dvipsnames,table]{xcolor}
\usepackage{booktabs}
\usepackage{capt-of}

\usepackage[numbers]{natbib}

\usepackage[
    colorlinks=true,
    linkcolor=MidnightBlue,
    citecolor=BrickRed,
    urlcolor=Blue,
]{hyperref}
\usepackage{doi}

\usepackage{xcolor}

\definecolor{cbBlue}{RGB}{0,114,178}
\definecolor{cbOrange}{RGB}{230,159,0}
\definecolor{cbGreen}{RGB}{0,158,115}
\definecolor{cbYellow}{RGB}{240,228,66}
\definecolor{cbRed}{RGB}{213,94,0}
\definecolor{cbPurple}{RGB}{204,121,167}
\definecolor{cbGray}{RGB}{187,187,187}

\definecolor{cbblue}{HTML}{0072B2}
\definecolor{cborange}{HTML}{D55E00}
\definecolor{cbred}{HTML}{CC79A7}

\newacronym{lino}{LiNO}{Lightweight Network for Occupancy Estimation}
\newacronym{pan}{PAN}{Pillars-Attention-based Network}
\newacronym{bev}{BEV}{Bird's-Eye View}
\newacronym{miou}{mIoU}{mean Intersection over Union}
\newacronym{iou}{IoU}{Intersection over Union}
\newacronym{rayiou}{RayIoU}{Ray-level Intersection over Union}
\newacronym{fps}{FPS}{frames per second}
\newacronym{vram}{VRAM}{video random-access memory}
\newacronym{mfa}{MFA}{Multi-modal Feature Aggregation}
\newacronym{mdca}{MDCA}{Multi-modal Deformable Cross-Attention}
\newacronym{rvt}{RVT}{Radar-assisted View Transformation}
\newacronym{lss}{LSS}{Lift-Splat-Shoot}
\newacronym{fpn}{FPN}{Feature Pyramid Network}
\newacronym{sota}{SOTA}{state-of-the-art}
\newacronym{gt}{GT}{ground truth}
\newacronym{fp16}{FP16}{16-bit floating point precision}
\newacronym{gpu}{GPU}{graphics processing unit}
\newacronym{cpu}{CPU}{central processing unit}
\newacronym{soc}{SoC}{system on chip}
\newacronym{crn}{CRN}{Camera Radar Net}
\newacronym{tpv}{TPV}{tri-perspective view}
\newacronym{lidar}{LiDAR}{Light Detection and Ranging}

\begin{document}

\title{LiNO: Lightweight Network for Occupancy Estimation}

\newcommand{\mi}{~(Member,~IEEE)}
\newcommand{\smi}{~(Senior Member,~IEEE)}

\author{
    Ruan~Bispo\mi,
    Tim~Brophy\mi,
    Reenu~Mohandas\mi,
    Anthony~Scanlan,
    Ciarán~Eising\smi%
    \thanks{
        The authors are with the Data Driven Computer Engineering Research
        Centre, Department of Electronic and Computer Engineering, University of
        Limerick, Limerick V94 T9PX, Ireland, and also with Lero, the Research
        Ireland Centre for Software, University of Limerick, Limerick V94 NYD3,
        Ireland. Corresponding author: Ruan Bispo
        (e-mail: bispodos.ruan@ul.ie).
    }%
    \thanks{
        This work was supported by Lero, the Research Ireland Centre for
        Software, under Grant 13/RC/2094\_P2.
    }%
    \thanks{
        This work has been submitted to the IEEE for
        possible publication. Copyright may be transferred
        without notice, after which this version may no longer be accessible.
    }%
    
}

\markboth{LiNO: Lightweight Network for Occupancy Estimation}{Bispo et al.}

\maketitle

\begin{abstract}
Occupancy estimation for automated vehicles is a crucial task that builds a 3D map using
vehicle sensors. This map allows the vehicle to identify drivable areas and various
obstacle classes. The task involves complex subtasks such as depth estimation, object
classification, and segmentation to align semantic information with the 3D grid.
Moreover, producing these estimations from multiple sensors demands significant
computational power and memory, as high-dimensional data fusion requires complex feature
extraction and processing. Consequently, most existing methods rely on large and
computationally intensive architectures that are not suitable for real-time or
real-world applications. We present a real-time, lightweight architecture that leverages
camera and radar sensors through a residual multi-stage head. Unlike prior methods
that rely on heavy camera-only blocks, our architecture employs camera-radar fusion and
lightweight components. Additionally, we introduce a task-specific data augmentation
strategy that enhances data diversity and improves generalisation. Our method achieves
\gls{sota} performance, attaining 45.7 \gls{miou} on Occ3D and 41.9 \gls{rayiou} (41.9) on OpenOcc datasets,
running at 17.2 \gls{fps} and using only 5.3 GB of memory, compared with
4.1 \gls{fps} and 7.7 GB for STCOcc.
\end{abstract}
\section{Introduction}
\label{sec:intro}

Road traffic safety remains a major societal challenge. In a U.S. crash
analysis conducted by the National Highway Traffic Safety Administration,
driver-related factors were identified as the critical reason in approximately
94\% of investigated crashes \cite{singh2015critical}. Although this does not
imply that human behaviour was the sole cause of these events, it highlights
the importance of vehicle systems capable of continuously monitoring the
environment and supporting timely hazard mitigation. Advanced driver-assistance
and automated driving systems can help mitigate some of these failures by
detecting hazards and supporting preventive actions. However, their safety
potential fundamentally depends on reliable environmental perception, since
missed obstacles, inaccurate free-space estimation, or incorrect semantic
interpretation may lead to unsafe vehicle behaviour
\cite{MUELLER2020310}.

Reliable perception of 3D environments is therefore a core requirement for automated vehicles, supporting safe navigation and informed decision-making. In perception tasks, 3D occupancy prediction aims to partition the surrounding space into voxels and assign semantic labels to each voxel, providing a detailed representation of drivable regions and obstacles. This dense spatial representation is beneficial for downstream applications such as mapping, planning, and motion prediction, all of which are crucial for robust autonomous operation \cite{sima2023sceneoccupancy, Xu_2025}.

In contrast to 3D bounding boxes that oversimplify object shapes, 3D occupancy provides a geometry-aware representation of the environment by encoding both object and background structures within a set of semantic cubes. While a 3D box can only capture the coarse shape of the objects, 3D semantic occupancy can recover fine-grained geometric details important for a complete perception system by leveraging multi-sensor information. This voxel-based representation enables a more precise reconstruction of objects. It simultaneously maps background classes such as roads, sidewalks, and vegetation, resulting in a richer and more comprehensive understanding of the 3D environment.

Despite recent advances, \gls{sota} methods still rely on computationally intensive and slow multi-frame-dependent temporal heads, as well as camera-only approaches, to improve accuracy \cite{liao2025stcocc,li2023fb,li2024bevformer,lin2024teocc}. This is largely due to the inherently high semantic dependence of the task. While effective, these
approaches incur high computational costs, increased latency, and substantial memory
usage, limiting their deployment in real-time or resource-constrained scenarios. A
deployment-efficient occupancy model must balance real-time performance, lightweight
design, and accuracy\cite{Xu_2025}.

In terms of memory available, despite advances in hardware, there is a critical
constraint for real-world autonomous platforms. For example, some versions of the Tesla
Model S Autopilot relied on the NVIDIA Parker \gls{soc}, which shares up to 8 GB of
memory between the \gls{cpu} and \gls{gpu} \cite{bateni2020co}. Even though more recent
platforms now support 16~GB or more of memory, such upgrades still come with higher
cost, power consumption, and thermal design requirements, and the overall stack must
still operate within stringent real-time and resource-limited budgets. Recent works
confirm that computational and memory demands continue to rise, imposing practical
deployment challenges in real vehicles \cite{wang2024movingforwardreviewautonomous}.

Another critical limitation in prior literature is the strong dependence on many
previous frames to achieve stable predictions (e.g., 8 or 16). These frames are
typically sampled as key frames over several seconds, introducing a temporal delay. For
example, methods aggregating 16 frames effectively operate with a delay of about 8
seconds, in addition to inference time, often below 5 \gls{fps}. At a driving speed of
50~km/h (assuming constant velocity), this corresponds to over 100 meters of vehicle
displacement before an updated occupancy estimation becomes available
\cite{bispo2025panpillarsattentionbasednetwork3d}. Furthermore, other studies on
automotive perception and control systems commonly adopt processing and rates around
10~\gls{fps} (or 100~ms) as a critical input for downstream modules, indicating that
operating below this rate may compromise timely reaction and overall safety
\cite{zhang2020optimal, hsiao2022zhuyi, kilian2022emergency, xu2022aoi}. For comparison,
\gls{lino} performs temporal fusion in \gls{bev} space, but only uses 3 previous BEV
feature sets, following \gls{crn} \cite{kim2023crncameraradarnet}.

Additionally, a fundamental challenge that must be addressed for a reliable 3D semantic
occupancy estimation lies in the supervision signal itself. Occupancy ground truth is
inherently sparse, with more than 90\% of voxels labelled as empty. Moreover, since
annotations are typically generated from annotated LiDAR point clouds, regions without
returns are often labelled as empty, even when valid objects may be present. This
results in false empty labels, noisy supervision, and incorrect labelling, further
complicating the learning process and biasing models toward predicting the right class
for each voxel. \cite{tang2024sparseocc}.

In this work, we challenge the assumption that computationally expensive architectures
are necessary for accurate occupancy estimation. We propose \gls{lino}, a lightweight
camera-radar architecture using a fusion-based \gls{bev} encoder and a residual dual
occupancy head. By directly modelling the semantic occupied state of 3D space without
relying on temporal multi-sensor aggregation, our method preserves spatial fidelity
without requiring resource-intensive temporal processing. The design simultaneously
addresses real-time latency, memory constraints, and fine-grained reconstruction,
achieving competitive accuracy while reducing both inference time
and memory footprint. Our main contributions are:
\begin{enumerate}
    \item We present \gls{lino}, an efficient camera-radar architecture for 3D
    semantic occupancy estimation that combines compact BEV fusion with lightweight
    volumetric decoding.
    \item We design a lightweight occupancy head for accurate real-time inference,
    consisting of a convolutional head and a U-Net residual fusion in a
    multi-stage dual-supervised approach.
    \item We introduce morphological closing as a data augmentation strategy for training
    to mitigate ground-truth sparsity and increase performance.
    \item LiNO achieves \gls{sota} \gls{miou} (45.7) on Occ3D and \gls{rayiou}
    (41.9) on OpenOcc, running at 17.2~\gls{fps} with 5.3~GB of memory compared
    with 4.1~\gls{fps} and 7.7~GB for STCOcc. Comprehensive ablations further
    validate the effectiveness of our architectural design.
\end{enumerate}
\section{Related Work}
\label{sec:rel_work}

\subsection{Camera-based Methods}
\label{subsec:cam_based}
Recent methods for 3D occupancy prediction mainly lift image features into voxel or
\gls{bev} space for semantic scene completion. Early approaches such as
MonoScene~\cite{cao2022monoscene}, TPVFormer~\cite{huang2023tri}, and
OccFormer~\cite{Zhang_2023_ICCV} employ dense volumetric encoders based on 3D
convolutions or transformers. MonoScene~\cite{cao2022monoscene} processes monocular
images through a 2D and 3D U-Net, while TPVFormer~\cite{huang2023tri} introduces
\gls{tpv} representations to capture geometric consistency across views.
OccFormer~\cite{Zhang_2023_ICCV} further enhances voxel encoding using a dual-path
transformer that jointly models spatial and semantic dependencies. Although these dense
volumetric designs achieve strong completion accuracy, their memory requirements and
extensive computation limit real-time deployment.

Early camera-based occupancy methods predominantly adopted dense voxel
representations because they provide a direct formulation for semantic scene
completion. By allocating features across the full 3D volume, these methods can
jointly model object geometry, free space, and background structures without
requiring a predefined sparsity prior \cite{cao2022monoscene,huang2023tri,Zhang_2023_ICCV}.
This design is particularly convenient in camera-only settings, where depth
ambiguity and occlusions make it difficult to determine in advance which 3D
regions should be processed. However, dense volumetric representations are
commonly coupled with computationally expensive 3D encoders and decoding heads,
as every region of the voxel grid must be processed, including the large fraction
corresponding to empty space. As a result, these architectures often incur high
memory consumption and latency, which limits their applicability in real-time
deployment scenarios \cite{cao2022monoscene,huang2023tri,Zhang_2023_ICCV,
tang2024sparseocc,mei2024camera}.

Other approaches explicitly rely on spatio-temporal occupancy estimation, such as
BEVFormer~\cite{li2024bevformer} and FB-Occ~\cite{li2023fb}, aggregate temporal context
using attention-based and forward-backward mechanisms, respectively, thereby improving
geometric consistency at the expense of memory and inference speed.
OccNet~\cite{sima2023sceneoccupancy} and BEVDet \cite{huang2022bevdet4d} also demonstrate
strategies for modelling long-term temporal dependencies in either the image feature or
3D voxel/BEV space. Compact transformer designs such as COTR~\cite{ma2024cotr} and
OPUS~\cite{wang2024opus} attempt to balance accuracy and efficiency through lightweight
attention modules, but still rely on temporal fusion and heavy backbones. Additionally,
STCOcc~\cite{liao2025stcocc} combines sparse spatial-temporal fusion with
occupancy-aware refinement, improving the \gls{rayiou} performance and memory footprint,
but it also has high latency.

Temporal fusion became popular because a single image frame provides incomplete
evidence of the 3D environment. Occluded regions, limited camera field of view,
and depth uncertainty can be partially resolved by aggregating observations
from previous timestamps, improving geometric consistency and scene completeness
\cite{li2024bevformer,li2023fb,huang2022bevdet4d,liao2025stcocc}. Nevertheless,
this improvement introduces a further deployment trade-off: multiple feature
maps must be stored, aligned, and fused across time, increasing memory usage and
inference latency. These limitations motivate occupancy architectures that retain
compact BEV-based spatial representations while using lightweight volumetric
decoding and avoiding dependence on long temporal contexts.

To address scalability and efficiency, several methods explore sparse 3D reasoning.
SparseOcc~\cite{tang2024sparseocc} introduces a fully sparse framework that restricts
computation to non-empty regions of the scene, while SGN~\cite{mei2024camera} adopts a
dense to sparse to dense strategy with adaptive voxel sampling and hybrid semantic
guidance. VoxFormer~\cite{li2023voxformer} proposes depth-based query proposals to guide
cross-attention between 2D image features and 3D voxels, effectively reducing redundant
computation. FlashOcc~\cite{yu2023flashocc} adopts a 2D approach instead of using 3D
convolutions, and it still suffers from accuracy limitations. Other sparse designs, such
as Symphonize~\cite{jiang2024symphonize}, leverage occupancy priors and semantic
filtering to refine scene reconstruction. Despite their efficiency gains, these methods
still suffer from substantial memory consumption or accuracy issues, which significantly
hinder their applicability in real-world applications.

\subsection{Multimodal Methods}
\label{subsec:multimodal_methods}

Multimodal fusion with radar remains relatively underexplored in 3D occupancy
estimation. Radar is an attractive complementary sensor because it provides
direct range measurements and motion-related cues, while remaining more robust
than cameras under adverse illumination and weather conditions. However, radar
measurements are sparse, noisy, and less semantically informative than image
features, which makes their association with visual observations and their
integration into dense 3D representations challenging \cite{kim2023crncameraradarnet,
lin2024teocc,wolters2025unleashinghydrahybridfusion, wang2025multiple}. Consequently, radar-based
occupancy systems must balance the geometric advantages of radar against the
additional complexity introduced by sensor calibration, feature alignment, and
fusion design.

TECOcc introduces temporal enhancement through camera-radar fusion to improve
occupancy prediction, but at the cost of increased computational requirements
\cite{lin2024teocc}. HyDRa~\cite{wolters2025unleashinghydrahybridfusion} extends
FB-Occ~\cite{li2023fb} with radar-camera temporal aggregation, showing promising
improvements in accuracy and robustness under adverse conditions. However,
HyDRa does not provide an inference-time or memory-usage analysis, and its
multimodal temporal fusion design is likely to introduce a higher computational
cost than the base FB-Occ model.

In this work, we focus on radar and camera inputs, as \gls{lidar} is typically used only
to generate ground-truth occupancy and depth labels for camera-to-BEV transformation and is not required during inference \cite{philion2020lift, liao2025stcocc, huang2022bevdet4d, li2023fb,
Xu_2025}. Using \gls{lidar} as an input would provide
substantially stronger geometric measurements and would therefore correspond to
a different sensor setting from the camera-radar scenario considered in this
work. Despite its accurate geometric sensing, LiDAR remains expensive for large-scale autonomous driving deployment, and mechanically rotating LiDARs introduce moving components that raise reliability and lifetime concerns under automotive vibration, shock, and temperature conditions \cite{yoo2018mems, li2020lidar}. 

This motivates the exploration of camera-radar alternatives for efficient 3D perception. Our goal is to develop a compact camera-radar architecture that combines BEV-level fusion with lightweight volumetric decoding for efficient occupancy estimation. Additionally, by avoiding heavy temporal fusion, our occupancy decoder remains agnostic to the temporal setup. It can operate on single-frame or temporally fused BEV inputs while maintaining computational efficiency. Finally, we compare our method using camera-only and camera-radar approaches to highlight the potential of camera-radar fusion as a scalable and deployment-friendly alternative for occupancy estimation.
\section{Methods}
\label{sec:methods}

\subsection{Overall Architecture}
\begin{figure*}[!t]
\centering
\includegraphics[width=1.0\linewidth]{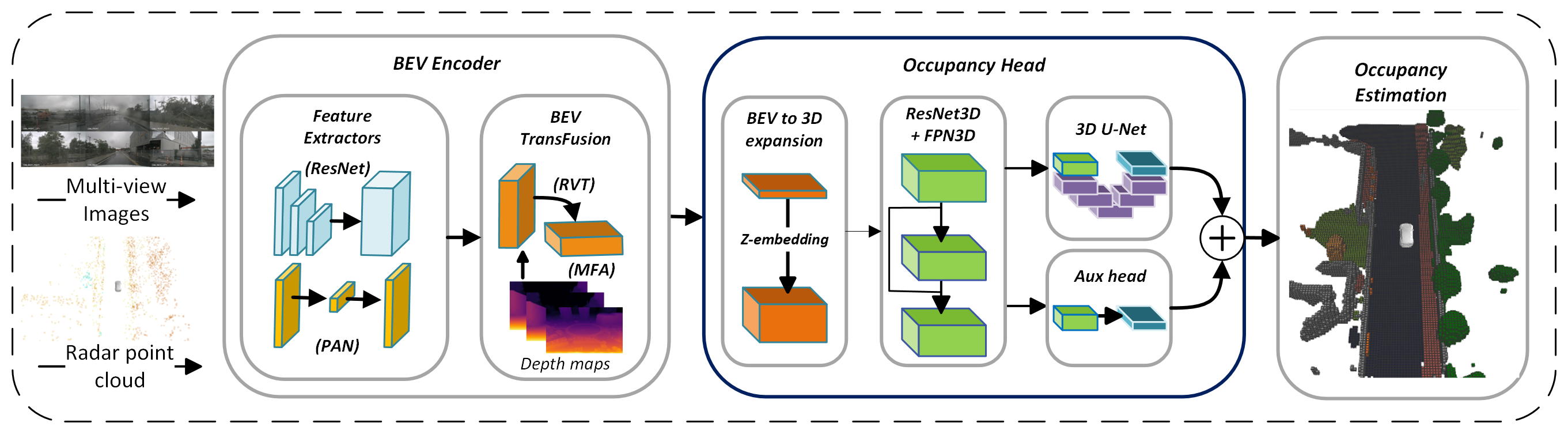}
\caption{
  Our proposed architecture using camera and radar data. Firstly, the camera
  and radar features are extracted using the respective feature extractors. Secondly,
  we lift and fuse these features into a BEV using a BEV Transformation and fusion
  module. Finally, the proposed occupancy head (highlighted in blue) processes the BEV
  features to produce the 3D occupancy estimation.
}
\label{fig_overall}
\end{figure*}

Our framework adopts a modular radar-camera fusion pipeline that transforms multi-sensor
inputs into a unified \gls{bev} representation for downstream 3D semantic occupancy
estimation. As illustrated in Fig.~\ref{fig_overall}, the system comprises a BEV encoder
and an occupancy head to compose the full architecture. The encoder handles feature
extraction from both camera and radar modalities, projects them into BEV space, and
performs multimodal fusion between camera and radar features in the BEV space. Our main
contribution lies in the design of this lightweight composition with a
dual-supervised occupancy head, detailed in Sec.~\ref{sec:head}, which enables efficient
and accurate dense occupancy prediction.

\subsection{BEV Encoder}
The BEV encoder utilises several lightweight and fast-inference modules for multi-modal
feature extraction and transformation, including \gls{pan}
\cite{bispo2025panpillarsattentionbasednetwork3d}, \gls{rvt} \cite{kim2023crncameraradarnet},
\gls{lss} projection \cite{huang2022bevdet4d}, and \gls{mfa} \cite{zhu2020deformable,
kim2023crncameraradarnet}. It processes six multi-view camera images and radar point
clouds from five sensors using independent backbones
\cite{he2015deepresiduallearningimage, bispo2025panpillarsattentionbasednetwork3d},
followed by a \gls{rvt} that projects the features into the BEV domain and a multi-modal
aggregation stage that efficiently fuses both modalities
\cite{kim2023crncameraradarnet}. Together, these components produce a unified BEV
feature map that serves as input to the occupancy head.

\subsubsection{Feature Extractors}
We adopt a ResNet backbone \cite{he2015deepresiduallearningimage} with a \gls{fpn}
\cite{lin2017feature} to extract features from the six camera views, balancing accuracy
and efficiency without the computational overhead of transformer-based architectures.
Radar features are obtained using the
\gls{pan}~\cite{bispo2025panpillarsattentionbasednetwork3d}, which extends the pillar
representation~\cite{lang2019pointpillarsfastencodersobject} with attention-based
refinement. The radar point cloud from five sensors is voxelized into a 2D grid where
only non-empty pillars are processed through an MLP encoder and a self-attention branch
that models long-range dependencies while reducing sparsity. A lightweight convolutional
block then restores spatial coherence, producing dense and geometrically grounded radar
embeddings that complement the camera features for subsequent fusion.

\subsubsection{BEV TransFusion}
We adopt the radar-assisted transformation and fusion pipeline proposed in
CRN~\cite{kim2023crncameraradarnet}, which combines two key modules. First, the
Radar-assisted View Transformation (RVT) extends LSS~\cite{philion2020lift} by
projecting image features into the BEV domain using radar-derived depth and occupancy
information, improving geometric accuracy. Second, the \gls{mfa} uses a \gls{mdca}
module to align and fuse radar and camera BEV features through adaptive sampling and
cross-modal deformable attention \cite{zhu2020deformable}, mitigating spatial
misalignment and calibration errors. The resulting unified BEV representation, enriched
with both semantic and geometric cues, is directly forwarded to the occupancy head for
dense prediction.

\subsection{Occupancy Head}
\label{sec:head}

Our proposed occupancy head was designed to efficiently decode BEV features into a
high-resolution 3D occupancy grid, using small and fast blocks. Unlike previous
volumetric decoders that rely on computationally demanding 3D convolutions or
multi-branch temporal fusion, our occupancy head employs a compact 3D architecture with
explicit vertical embedding, hierarchical volumetric reasoning, and dual-supervised
learning. This design enables precise spatial understanding and stable convergence at
low computational cost. The overall architecture consists of five main components: a
learnable height embedding module ($Z$-embedding), a 3D BEV backbone (ResNet3D), a
feature pyramid neck (FPN3D), an auxiliary head, and a 3D U-Net decoder for
voxel-wise prediction, as illustrated in Fig.~\ref{fig:occ_head}.

\subsubsection{Volumetric Expansion and $Z$-Embedding}
Given the fused BEV features from the encoder, we first expand them along the vertical
dimension, repeating the features to form a volumetric representation. This 2D-to-3D
expansion allows the model to reason over height-dependent occupancy patterns such as
elevated structures or obstacles. To enrich this representation, we introduce a
learnable $Z$-Embedding module that assigns a distinct channel-wise bias to each
vertical level, allowing the network to encode altitude-specific semantics without
requiring explicit geometric priors. This allows the network to differentiate features
at different vertical levels:
\begin{equation}
\mathbf{V}_z = \mathbf{V}_0 + \mathbf{e}_z, \quad \text{for } z = 1, \dots, Z
\label{eq:z_embed}
\end{equation}
where $\mathbf{V}_0$ is the expanded BEV features and $\mathbf{e}_z$ is a learnable
embedding vector for the $z$-th height bin and the final $\mathbf{V}_z$ represents the
volumetric features. For this task, we used $Z=16$, representing the 3D discretised
height.

\subsubsection{3D Feature Encoding}
The embedded volumetric features are then processed by a ResNet3D backbone composed of
three residual stages with strides $\{1,2,2\}$ and channel widths $\{64,96,128\}$. This
hierarchical encoder captures volumetric context while maintaining efficiency through 3D
convolutions and batch normalisation. To recover fine spatial details lost during
downsampling, we employ a 3D Feature Pyramid Network (FPN3D) as the neck. The FPN
aggregates multi-scale features from all backbone stages through lateral and top-down
connections, producing a uniform-resolution feature map with 96 channels. This design
enhances geometric consistency across height levels and improves the representation of
thin or partially observed structures, which are especially challenging under sparse
radar–camera input.

\begin{figure*}[t]  
\centering
\begin{minipage}{0.47\linewidth}
  \centering
  \includegraphics[width=\linewidth]{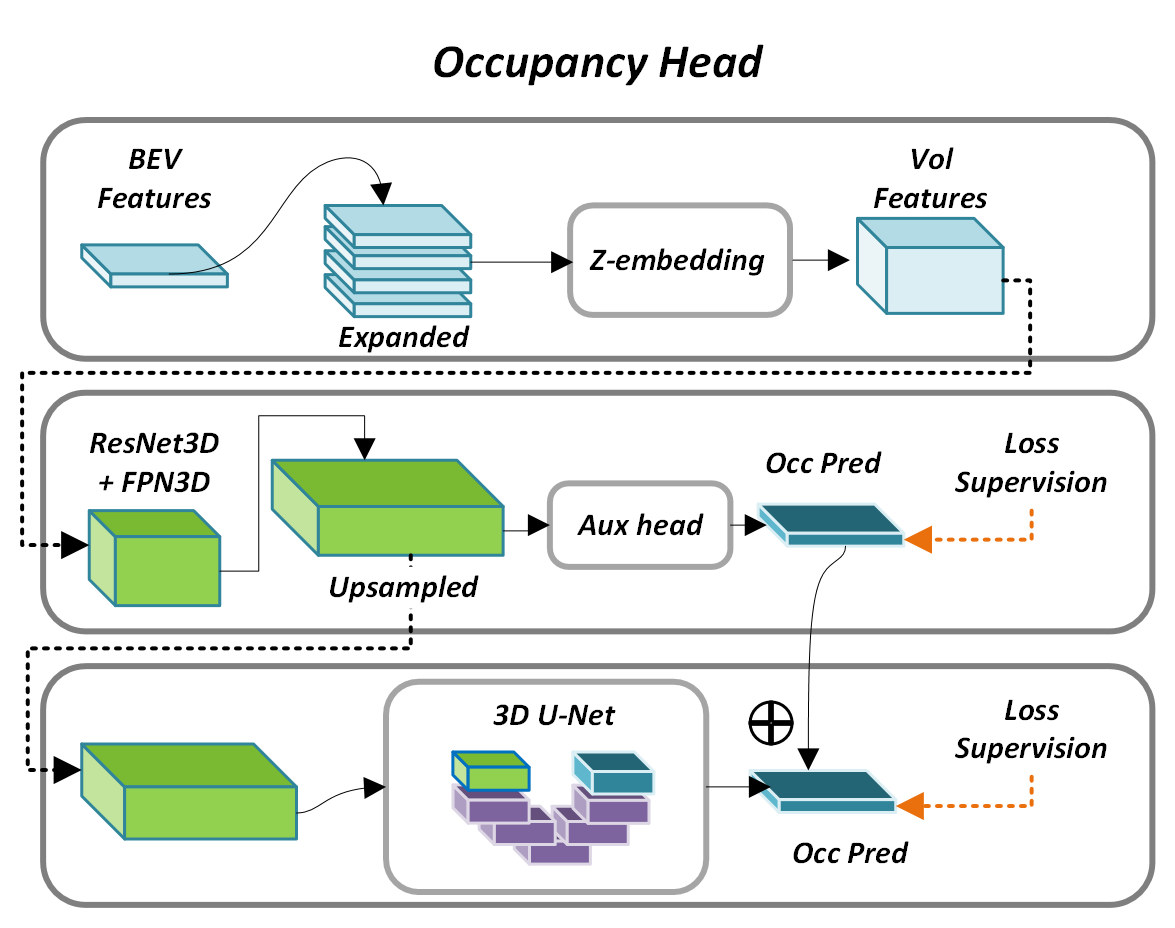}
  \captionof{figure}{Overview of the proposed dual-supervised Occupancy Head.
  The BEV features are expanded and enriched via $Z$-embedding, which are processed by a ResNet3D backbone and an FPN3D neck.
  An auxiliary head and a final 3D U-Net decoder perform hierarchical decoding with independent supervision.
  The final occupancy prediction is obtained by residual fusion of both predictions.}
  \label{fig:occ_head}
\end{minipage}%
\hspace{0.03\linewidth}  
\begin{minipage}{0.47\linewidth}
  \centering
  \includegraphics[width=\linewidth]{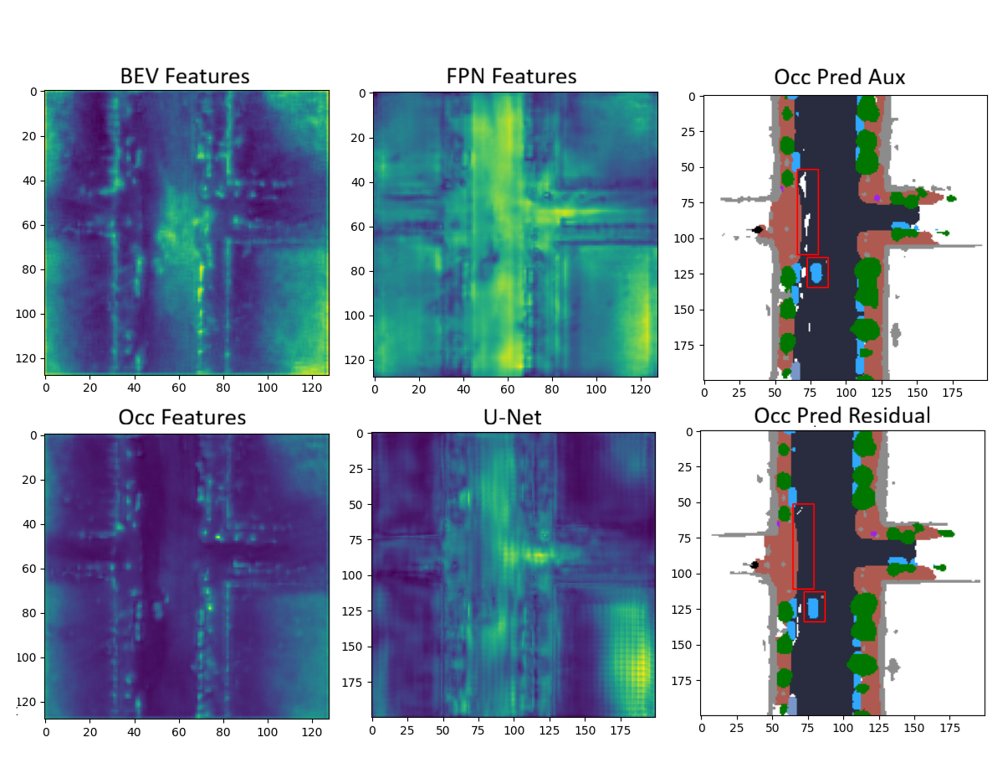}
  \captionof{figure}{
  Visualisation of top-view feature maps across different stages of the proposed head.
  The BEV, FPN, and Occ features correspond to the inputs of the three main blocks illustrated in Fig.~\ref{fig:occ_head}, while the U-Net output represents the final stage before prediction.
  The two rightmost maps show occupancy predictions from the auxiliary head and after residual fusion, respectively.
  }
  \label{fig:occ_feat_map}
\end{minipage}
\end{figure*}

\subsubsection{Occupancy Decoding}
The unified volumetric features are upsampled using trilinear interpolation to match the
target voxel resolution $(X, Y, Z)$ before decoding. This operation restores the spatial
scale of the feature tensor along the horizontal and vertical axes, ensuring alignment
between the BEV representation and the final prediction grid. The upsampled features are
then decoded by a 3D U-Net \cite{ronneberger2015u} head configured with channel stages
$\{32, 64, 128\}$, following a symmetric encoder-decoder design with skip connections
that preserve structural continuity and fine boundary details. An auxiliary
convolutional head is attached before the final decoder, consisting of a single 3D
convolutional layer. This auxiliary head is also supervised and serves as a shallow
refinement stage that enhances local compactness and stabilises feature learning before
the final decoding \cite{he2015deepresiduallearningimage}.

\subsubsection{Dual-Supervised Training}
To promote hierarchical feature learning, both the auxiliary head and the final 3D U-Net
output are supervised using the same occupancy labels but with different weights.
Specifically, the auxiliary prediction $\mathcal{H}_{aux}$ receives a reduced
supervision weight $\alpha < 1$, while the final 3D U-Net prediction $\mathcal{H}_{U}$
is trained with full loss weight. Following prior works on the deep supervision field
\cite{lee2015deeply, xie2015holistically}, we expanded the concept to our
dual-supervised head, which yielded stable convergence in practice. The dual supervision
objective is formulated as:
\begin{equation}
\mathcal{L}_{dual} = \alpha \, \mathcal{L}_{occ}(\mathcal{H}_{aux}) + \mathcal{L}_{occ}(\mathcal{H}_{U})
\label{eq:dual_loss}
\end{equation}
where $\mathcal{L}_{occ}$ denotes the multi-term occupancy loss defined in
Sec.~\ref{sec:loss}. This dual supervision enforces consistent optimisation across
hierarchical depths of the decoder, alleviating vanishing gradients, improving gradient
flow, and convergence stability, particularly during early training stages
\cite{xie2015holistically, szegedy2015going, lee2015deeply}.

The final volumetric output is computed through residual fusion of both predictions:
\begin{equation}
\mathbf{Y} = \mathcal{H}_{U}(\mathbf{F}_{\text{FPN}}) + \mathcal{H}_{aux}(\mathbf{F}_{\text{FPN}})
\label{eq:residual_fusion}
\end{equation}
producing per-voxel class logits $\mathbf{Y} \in \mathbb{R}^{B \times C \times Z \times
X \times Y}$ over a fixed grid of $200 \times 200 \times 16$ voxels, where
${B,~C,~Z,~X,~Y}$ are the batch size, number of channels, and $(Z, X, Y)$ correspond to
the spatial axes, respectively. The resulting visualisation of our head is shown in
Fig.~\ref{fig:occ_feat_map}, illustrating the progression of the feature space across
stages and highlighting the benefit of the multi-stage approach for improving the
completeness of the final estimation.

This head design effectively balances representational richness, stability, and runtime
efficiency. The combination of learnable height embeddings, hierarchical 3D convolution,
and dual-supervised learning enables accurate and robust occupancy estimation even under
strong sparsity or sensor noise. In practice, our head achieves real-time inference and
demonstrates improved geometric continuity compared to heavier transformer-based
decoders, validating its suitability for practical 3D perception systems.

\subsection{Loss Formulation}
\label{sec:loss}
We compute the occupancy loss for each decoding occupancy prediction, combining
voxel-wise and structural objectives to ensure both semantic accuracy and geometric
consistency. Following common practices in 3D occupancy
prediction~\cite{cao2022monoscene, li2023fb, liao2025stcocc}, we adopt a multi-term
formulation that jointly optimises class balance, volumetric continuity, and physical
plausibility.

Given the inherent sparsity and imbalance of 3D scenes, we employ a weighted Focal Loss
($\mathcal{L}_{focal}$)~\cite{lin2017focal} to emphasize under-represented classes,
together with the Lovasz-Softmax ($\mathcal{L}_{lovaz}$)~\cite{berman2018lovasz} and
Dice loss ($\mathcal{L}_{dice}$)~\cite{li2019dice} to improve boundary precision and
volumetric overlap. Additionally, we integrate the Scene-Class Affinity losses, proposed
by~\cite{cao2022monoscene}, namely the Semantic-Scaling ($\mathcal{L}_{sem\_scal}$) and
Geometric-Scaling ($\mathcal{L}_{geo\_scal}$) terms, which encourage consistency between
the semantic and geometric structure of predicted scenes. The overall occupancy loss for
each stage $i$ is defined as:
\begin{equation}
\mathcal{L}^i_{occ} =
\mathcal{L}_{focal} +
\mathcal{L}_{dice} +
\mathcal{L}_{lovasz} +
\mathcal{L}_{sem\_scal} +
\mathcal{L}_{geo\_scal}
\label{eq:occ_loss}
\end{equation}

In addition to occupancy estimation, the model jointly learns a depth prediction
objective, supervised by a Cross-Entropy Loss~\cite{zhang2018generalized} applied to the
voxelized depth distribution, following the formulation of LSS~\cite{philion2020lift}.
The total optimisation objective is therefore expressed as:
\begin{equation}
\mathcal{L} =
\lambda_d \mathcal{L}_{depth} +
\sum_{i=1}^{N} w_i \mathcal{L}^i_{occ}
\label{eq:total_loss}
\end{equation}
where $w_i$ denotes the supervision weight for the $i$-th stage in the dual supervised
pathway, and $\lambda_d$ controls the contribution of the depth regularisation term.
Following prior works~\cite{cao2022monoscene, li2023fb}, we expand the default balancing
strategy between losses in a multi-loss approach, which provides stable convergence
without additional tuning. This combination of semantic, geometric, and depth objectives
promotes robust 3D reasoning and ensures accurate volumetric reconstruction even under
severe class imbalance or sparse observations~\cite{li2023fb}.

\subsection{Class-Wise Morphological Closing for GT Completion}
\label{sec:gt_da}

To mitigate sparsity in the occupancy labels, we apply a class-wise 3D morphological
closing operation during training. Each class is processed separately using a kernel
adapted to its geometry: flat classes (e.g., roads, terrain) use a planar kernel to fill
empty voxels, while non-flat classes (e.g., cars, buses) use a cubic kernel. This
approach fills empty voxels, increasing the completeness of the ground truth while
preserving object boundaries. Let $L_c$ denote the set of voxels belonging to class $c$.
The morphological closing of each class is defined as
\begin{equation}
\hat{L}_c = (L_c \oplus S_c) \ominus S_c, \quad c \in \mathcal{C} ,\ c \neq \text{empty}
\end{equation}
where $\oplus$ and $\ominus$ denote 3D dilation and erosion, $\mathcal{C}$ is the set of
semantic classes, and $S_c$ is the kernel applied to class $c$, defined as
\begin{equation}
S_c =
\begin{cases}
S_\text{flat}, & c \in \mathcal{C}_\text{flat},\\
S_\text{non-flat}, & c \in \mathcal{C}_\text{non-flat}.
\end{cases}
\end{equation}
Here, $S_\text{flat}$ is a planar kernel ($5 \times 5 \times 1$) applied to flat
classes, and $S_\text{non-flat}$ is a cubic kernel ($3 \times 3 \times 3$) for non-flat
classes, where further details on the kernel selection are provided in the ablation
study Sec. \ref{sec:ablation}, Tab.~\ref{tab:mask_aug}. Finally, the completed ground-truth label map is
obtained by combining all processed classes:
\begin{equation}
\hat{L} = \bigcup_{c \in \mathcal{C},\ c \neq \text{empty}} \hat{L}_c
\end{equation}
\section{Experiments}
\label{experiments}
\subsection{Dataset and Metrics}
We evaluate our approach on the Occ3D~\cite{tian2023occ3d} and
OpenOcc~\cite{sima2023sceneoccupancy} benchmarks derived from the nuScenes
dataset~\cite{caesar2020nuscenes}, which provides voxelized 3D semantic occupancy
labels. The datasets comprise 600 outdoor driving scenes for training, 150 for
validation, and 150 for testing. Each scene has approximately a 20~s duration, and the
key samples are annotated every 0.5~s. Following standard practice~\cite{liao2025stcocc,
li2023fb, li2024bevformer, tang2024sparseocc, wang2024opus, ma2024cotr,
huang2022bevdet4d, pan2024renderocc}, all quantitative results are reported on the
validation set, since test annotations are not publicly available and all prior works
use the same split for comparison.

The Occ3D dataset, our main benchmark, has a volume size of \([200, 200, 16]\), voxel
size of \([0.4m, 0.4m, 0.4m]\), occupancy range of \([80m, 80m, 6.4m]\), and 18 classes,
including others, barrier, bicycle, bus, car, construction vehicle, motorcycle,
pedestrian, traffic cone, trailer, truck, drivable surface, other flat, sidewalk,
terrain, manmade, vegetation, and empty. This dataset exhibits strong class imbalance,
as more than 90\% of the voxels belong to the empty class \cite{tang2024sparseocc}. We
also evaluate on OpenOcc~\cite{sima2023sceneoccupancy} as a second benchmark protocol.
OpenOcc is derived from nuScenes but uses a different occupancy annotation setup, with a 
denser voxel distribution and 17 semantic classes, excluding only the label
``others''. Reporting both benchmarks allows us to confirm generalisation and whether the observed gains hold
beyond a single occupancy annotation convention.

There are two primary metrics for occupancy evaluation: mean Intersection over Union
(mIoU) and Ray-based Intersection over Union (RayIoU). The literature does not establish
a clear preference between them, as some works report RayIoU while others rely on mIoU.
Prior studies have noted that mIoU may be sensitive to overestimation of surface
thickness. At the same time, both metrics exhibit known limitations related to empty
voxels and false positives \cite{tang2024sparseocc}. To ensure comprehensive and fair
comparison with existing methods, we report results using both metrics, consistent with
the metrics reported in prior work.

\begin{figure}[!t]
\centering
\includegraphics[width=\columnwidth,height=0.3\textheight,keepaspectratio]{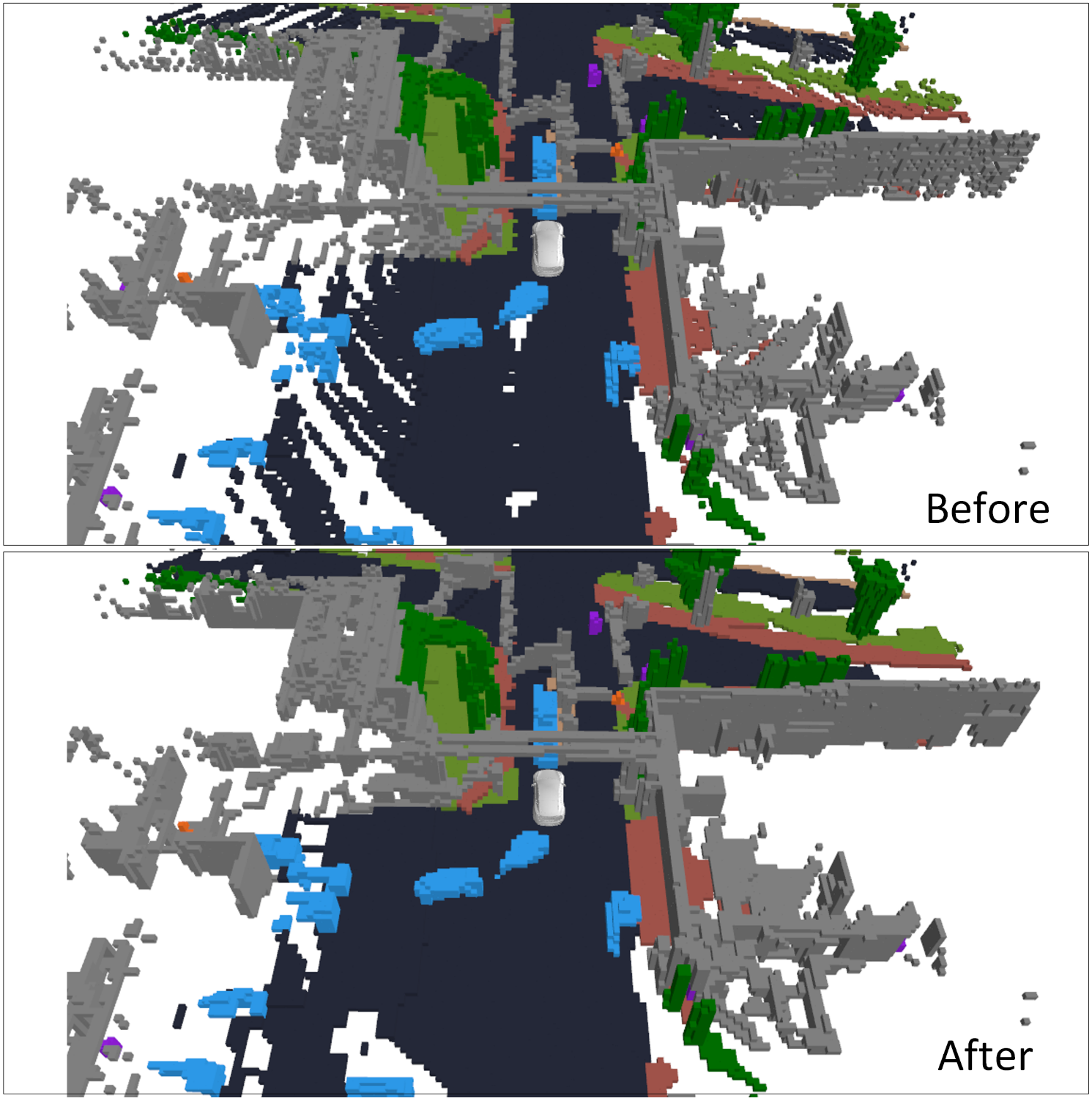}
\caption{Proposed data augmentation based on morphological closing to fill empty regions.}
\label{fig:aug_example}
\end{figure}

\subsection{Implementation Details}
We implement our model using PyTorch~\cite{paszke2019pytorch}. Following previous works,
we adopt ResNet-50 as the image backbone for our base model, and ResNet-18 as the image
backbone to compose a faster version. Each baseline is evaluated using the
hyperparameters and configurations originally optimised for its own setup, as these
methods rely on distinct camera layouts, image resolutions, and training schedules. Our
model is trained for 90 epochs with a global batch size of 8, using the AdamW
optimizer~\cite{kingma2017adammethodstochasticoptimization} in an end-to-end manner. The
baseline learning rate is set to 2e-4 and reduced by a factor of 0.1 in the last five
and final epochs. For a fair comparison, we accumulate three previous BEV feature maps
in the BEV encoder, following a similar strategy to~\cite{kim2023crncameraradarnet}. All
comparative results were obtained on a NVIDIA A100 GPU with a batch size of 1 and
\gls{fp16} precision using PyTorch to ensure a fair comparison.

To improve spatial consistency in the voxelised labels, we are applying the proposed
morphological closing operation as data augmentation, as mentioned in Section
\ref{sec:gt_da}, with a probability of 0.2. The closing is applied to the \gls{gt}
occupancy map during training, as illustrated for the example in
Fig.~\ref{fig:aug_example}.

\subsection{Results}

The quantitative comparisons are reported in Tab.~\ref{tab:occ3d_miou},
Tab.~\ref{tab:results}, and Tab.~\ref{tab:openocc_results}. Tables~\ref{tab:occ3d_miou}
and~\ref{tab:results} evaluate Occ3D using mIoU and RayIoU, respectively,
whereas Tab.~\ref{tab:openocc_results} follows the established OpenOcc protocol
and reports RayIoU, consistent with recent methods such as STCOcc
\cite{liao2025stcocc}. Since there is no established consensus on a single
evaluation metric for occupancy prediction, we report results across datasets
and their corresponding evaluation protocols rather than relying on a single
leaderboard.

Importantly, LiNO achieves the best performance in two out of the three benchmark settings and remains
highly competitive in the third, demonstrating consistent accuracy-efficiency behaviour across evaluation criteria. These results are achieved with a substantially more efficient architecture that runs at 17.2~\gls{fps} with 5.3~GB of memory, while avoiding the heavy temporal fusion and computationally expensive decoding heads commonly used by high-performing occupancy methods. This demonstrates that competitive semantic occupancy estimation does not necessarily require large temporal contexts or high computational costs.

\begin{table*}[!t]
\centering
\small
\caption{
  Comparison of mIoU performance on the Occ3D~\cite{tian2023occ3d} dataset.
* indicates models trained with camera masks.
\textbf{M} denotes the input modality: C for camera, R for radar, and CR for both.
Higher values of IoU indicate better performance.}
\setlength{\tabcolsep}{2pt}
\resizebox{\textwidth}{!}{
\begin{tabular}{l|ccc|c|ccccccccccccccccc}
\toprule
\textbf{Method} &
\textbf{Backb.} &
\textbf{Input} &
\textbf{M} &
\textbf{mIoU $\uparrow$} &
\rotatebox{90}{\textbf{others}} &
\rotatebox{90}{\textbf{barrier}} &
\rotatebox{90}{\textbf{bike}} &
\rotatebox{90}{\textbf{bus}} &
\rotatebox{90}{\textbf{car}} &
\rotatebox{90}{\textbf{constrv}} &
\rotatebox{90}{\textbf{mcycle}} &
\rotatebox{90}{\textbf{ped}} &
\rotatebox{90}{\textbf{tcone}} &
\rotatebox{90}{\textbf{trailer}} &
\rotatebox{90}{\textbf{truck}} &
\rotatebox{90}{\textbf{drivsurf}} &
\rotatebox{90}{\textbf{othflat}} &
\rotatebox{90}{\textbf{sidewalk}} &
\rotatebox{90}{\textbf{terrain}} &
\rotatebox{90}{\textbf{manmade}} &
\rotatebox{90}{\textbf{veg}} \\
\midrule

BEVFormer \cite{li2024bevformer} & R101 & 1600×900 & C & 26.9 &
5.9 & 37.8 & 17.9 & 40.4 & 42.4 & 7.4 & 23.9 & 21.8 & 21.0 & 22.4 & 30.7 & 55.4 & 28.4 & 36.0 & 28.1 & 20.0 & 17.7 \\

CTF-Occ \cite{tian2023occ3d} & R101 & 1600×900 & C & 28.5 &
8.1 & 39.3 & 20.6 & 38.3 & 42.2 & 16.9 & 24.5 & 22.7 & 21.1 & 23.0 & 31.1 & 53.3 & 33.8 & 38.0 & 33.2 & 20.8 & 18.0 \\

TPVFormer \cite{huang2023tri} & R101 & 1600×900 & C & 27.8 &
7.2 & 38.9 & 13.7 & 40.8 & 45.9 & 17.2 & 20.0 & 18.9 & 14.3 & 26.7 & 34.2 & 55.7 & 35.5 & 37.6 & 30.7 & 19.4 & 16.8 \\

OSP* \cite{shi2024occupancy} & R101 & 1600×900 & C & 41.2 &
10.9 & 49.0 & 27.7 & 50.2 & 55.9 & 22.9 & 31.0 & 30.9 & 30.3 & 35.6 & 31.2 & 82.1 & 42.6 & 51.9 & 55.1 & 44.8 & 38.2 \\

SparseOcc \cite{tang2024sparseocc} & R50 & 704×256 & C & 30.9 &
10.6 & 39.2 & 20.2 & 32.9 & 43.3 & 19.4 & 23.8 & 23.4 & 29.3 & 21.4 & 29.3 & 67.7 & 36.3 & 44.6 & 40.9 & 22.0 & 21.9 \\

FlashOcc(M5)* \cite{yu2023flashocc} & R50 & 704×256 & C & 38.8 & 11.0 & 47.2 & 22.2 & 41.4 & 50.6 & 23.2 & 23.0 & 23.8 & 26.2 & 32.1 & 37.4 & 82.2 & 42.5 & 53.3 & 56.6 & 46.7 & 39.7 \\

FB-Occ* \cite{li2023fb} & R50 & 704×256 & C & 39.1 &
13.6 & 44.7 & 27.0 & 45.4 & 49.1 & 25.2 & 26.3 & 27.9 & 27.8 & 32.3 & 36.8 & 80.1 & 42.8 & 51.2 & 55.1 & 42.2 & 37.5 \\

ViewFormer* \cite{li2024viewformer} & R50 & 704×256 & C & 41.9 &
12.9 & 50.1 & 27.9 & 44.6 & 52.9 & 22.4 & 29.6 & 28.0 & 29.2 & 35.2 & 39.4 & \textbf{84.7} & \textbf{49.4} & 57.4 & 59.7 & 47.4 & 40.6 \\

COTR* \cite{ma2024cotr} & R50 & 704×256 & C & 44.5 &
13.3 & 52.1 & 31.9 & 46.0 & 55.6 & \textbf{32.6} & 32.8 & 30.4 & 34.1 & 37.7 & 41.8 & 84.5 & 46.2 & \textbf{57.6} & \textbf{60.7} & 51.9 & \textbf{46.3} \\

STCOcc* \cite{liao2025stcocc} & R50 & 704×256 & C & 44.6 &
\textbf{15.3} & \textbf{52.9} & 31.6 & 46.4 & 55.9 & 31.5 & 32.6 & 32.1 & 34.5 & 39.5 & 42.5 & 83.6 & 47.8 & 56.4 & 60.1 & 50.8 & 44.8 \\

\midrule
TECOcc* \cite{lin2024teocc} & R50 & 704×256 & CR & 42.9 & 10.8 & 50.3 & 24.3 & 49.0 & 57.3 & 29.4 & 24.4 & 30.1 & 28.5 & 36.5 & 43.0 & 84.0 & 43.1 & 56.0 & 59.3 & 54.2 & 49.2 \\

HyDRa* \cite{wolters2025unleashinghydrahybridfusion} & R50 & 704×256 & CR & 44.4 &
15.1 & 51.1 & 32.7 & 52.3 & 56.3 & 29.4 & \textbf{35.9} & 35.1 & 33.7 & 39.1 & 44.1 & 80.4 & 45.1 & 52.0 & 55.3 & \textbf{52.1} & 44.4 \\

\rowcolor{blue!10}
\textbf{LiNO* (ours)} & R50 & 704×256 & CR & \textbf{45.7} &
14.5 & 52.6 & \textbf{35.0} & \textbf{52.5} & \textbf{60.5} & 31.6 & \textbf{35.9} & \textbf{38.0} & \textbf{36.5} & \textbf{41.5} & \textbf{47.2} & 82.0 & 43.3 & 53.7 & 56.0 & 51.7 & 44.4 \\

\bottomrule
\end{tabular}
}
\label{tab:occ3d_miou}
\end{table*}

\begin{table*}[htbp]
\centering
\caption{
Comparison of RayIoU performance on the Occ3D~\cite{tian2023occ3d} dataset. We report
RayIoU at multiple distances (1m, 2m, 4m, and overall), along with \gls{vram} usage and
runtime in \gls{fps}. All methods were evaluated using an A100 GPU. For TEOcc-RC VRAM, 
the train code was not available to measure, and there are no reported metrics in the original paper.
Unavailable for replication or unreported values are marked with a dash (-).
}
\renewcommand{\arraystretch}{1.0}
\label{tab:results}
\resizebox{\textwidth}{!}{%
\begin{tabular}{
    l 
  | c 
    c 
  | c 
  | c 
    c 
    c 
  | c 
  | c 
   c 
}
\toprule
\textbf{Method} &
\textbf{Backb.} &
\textbf{Input} &
\textbf{Venue} &
\textbf{RayIoU\textsubscript{1m}}$\uparrow$ &
\textbf{RayIoU\textsubscript{2m}}$\uparrow$ &
\textbf{RayIoU\textsubscript{4m}}$\uparrow$ &
\textbf{RayIoU}$\uparrow$ &
\textbf{VRAM}$\downarrow$ &
\textbf{FPS}$\uparrow$ \\
\midrule
BEVFormer (4f) \cite{li2024bevformer} & R101   & 1600×900  & PAMI 2024 & 26.1 & 32.9 & 38.0 & 32.4 & 24.1 GB & 3.0  \\
RenderOcc (6f) \cite{pan2024renderocc} & Swin-B & 1408×512 & ICRA 2024 & 13.4 & 19.6 & 25.5 & 19.5 & 17.5 GB & -  \\
BEVDet-Occ (8f) \cite{huang2022bevdet4d} & R50 & 704×384 & arXiv 2022 & 26.6 & 33.1 & 38.2 & 32.6 & 10.2 GB & 2.6 \\
FB-Occ (16f) \cite{li2023fb} & R50 & 704×256 & ICCV 2023 & 26.7 & 34.1 & 39.7 & 33.5 &  9.6 GB & 10.3 \\
TEOcc-RC (9f)~\cite{lin2024teocc} & R50 & 704×256 & arXiv 2024 & 32.0 & 36.2 & 41.2 & 35.8 & - & 1.1 \\
SparseOcc (16f) \cite{tang2024sparseocc} & R50 & 704×256 & ECCV 2024 & 30.2 & 36.8 & 41.2 & 36.1 & 23.7 GB & 12.5 \\
COTR (8f) ~\cite{ma2024cotr} & R50  & 704×256 & NeurIPS 2024 & 36.3 & 41.7 & 45.1 & 41.0 &  18.3 GB & - \\
OPUS-L (8f) ~\cite{wang2024opus} & R50  & 704×256 & NeurIPS 2024 & 34.7 & 42.1 & 46.7 & 41.2 &  12.1 GB & 7.2 \\
STCOcc (16f) ~\cite{liao2025stcocc} & R50  & 704×256 & CVPR 2025 & 36.2 & 42.7 & 46.4 & \textbf{41.7} &  7.7 GB & 4.1 \\
\midrule
\rowcolor{blue!10}
\textbf{LiNO-cam-only (ours)} & R50    & 704×256   & - & 29.7 & 37.1 & 41.9 & 36.2 &  5.1 GB & \textbf{20.6}\\
\rowcolor{blue!10}
\textbf{LiNO-small (ours)} & R18  & 704×256   & - & 33.4 & 39.3 & 43.4 & 38.7 &  \textbf{4.7} GB & 19.0   \\
\rowcolor{blue!10}
\textbf{LiNO (ours)} & R50    & 704×256   & - & 35.7 & 42.0 & 45.9 & 41.2 &  5.3 GB & 17.2 \\
\bottomrule
\end{tabular}
}
\end{table*}

On Occ3D (Tab.~\ref{tab:occ3d_miou}), LiNO establishes a new SOTA in mIoU, outperforming
all prior methods, including recent camera-only and camera-radar approaches. Notably,
LiNO surpasses the last SOTA models such as STCOcc \cite{liao2025stcocc} and COTR
\cite{ma2024cotr} despite relying on compact BEV representations and without employing
deep temporal heads. Under the RayIoU protocol on Occ3D \cite{tian2023occ3d}
(Tab.~\ref{tab:results}), LiNO achieves 41.2\%, matching the second best-performing
method OPUS \cite{wang2024opus}. While slightly below the top score, LiNO runs at
17.2 FPS (STCOcc: 4.1 FPS) and requires only 5.3 GB of memory (STCOcc: 7.7 GB),
highlighting a substantially better accuracy-efficiency trade-off. Competing methods that achieve
marginally higher RayIoU rely on heavy temporal aggregation over up to 16 frames, whereas
LiNO operates without deep temporal fusion, using a lightweight multimodal design. The
profiled TEOcc-RC baseline is substantially slower under the same inference setup,
despite using a radar-camera input.
The camera-only variant in Tab.~\ref{tab:results} isolates the contribution of radar under
the same architecture. Adding radar improves overall RayIoU from 36.2 to 41.2,
indicating that radar provides complementary geometric evidence beyond visual features.
In particular, radar-derived range cues help reduce depth ambiguity in the BEV
representation, while Doppler measurements provide motion-sensitive information at low
sensing cost.

On OpenOcc (Tab.~\ref{tab:openocc_results}), we follow the evaluation protocol
adopted by recent OpenOcc methods, including STCOcc~\cite{liao2025stcocc}, and
therefore report RayIoU rather than mIoU. LiNO achieves SOTA
RayIoU performance on this second benchmark, surpassing STCOcc and other recent
approaches. Since OpenOcc follows a different occupancy annotation protocol
from Occ3D, with denser occupied regions and different label-noise
characteristics, its results provide complementary evidence that LiNO is not
tuned only to the Occ3D label distribution. Together, the Occ3D mIoU/RayIoU and
OpenOcc RayIoU results show that, while no single metric fully captures
occupancy quality, LiNO consistently ranks at or near the top across datasets
and evaluation protocols while retaining real-time inference.

\begin{table}[!t]
\centering
\small
\caption{
  Comparison of RayIoU performance on OpenOcc~\cite{sima2023sceneoccupancy},
  our second benchmark. OpenOcc uses a different occupancy annotation protocol from
  Occ3D and contains denser voxel occupancy. LiNO achieves SOTA performance,
  demonstrating generalisation beyond a single dataset convention. $^\dagger$ denotes
  results reported in \cite{liao2025stcocc}.
}
\label{tab:openocc_results}
\setlength{\tabcolsep}{5pt}
\begin{tabular}{@{}l|cc|c@{}}
\toprule
\textbf{Method} & \textbf{Backb.} & \textbf{Input} & \textbf{RayIoU}$\uparrow$ \\
\midrule
OccNeRF-C~\cite{zhang2023occnerf} & R101 & 1600$\times$900 & 21.6 \\
RenderOcc~\cite{pan2024renderocc} & R101 & 1600$\times$900 & 36.7 \\
OccNet~\cite{sima2023sceneoccupancy} & R101 & 1600$\times$900 & 39.7 \\
BEVFormer$^\dagger$~\cite{li2024bevformer} & R50 & 1600$\times$900 & 28.1 \\
FB-Occ$^\dagger$~\cite{li2023fb} & R50 & 704$\times$256 & 32.3 \\
SparseOcc$^\dagger$~\cite{tang2024sparseocc} & R50 & 704$\times$256 & 33.4 \\
STCOcc~\cite{liao2025stcocc} & R50 & 704$\times$256 & 40.8 \\
\midrule
\rowcolor{blue!10}
\textbf{LiNO (ours)} & R50 & 704$\times$256 & \textbf{41.9} \\
\bottomrule
\end{tabular}%
\end{table}

\subsection{Ablation Study}
\label{sec:ablation}

To evaluate the impact of each component in our model, we perform ablations on
Occ3D~\cite{tian2023occ3d}, reporting overall RayIoU under a reduced setup with 30\% of
the full dataset, 24 epochs, and using the ResNet-18 version. The ablations isolate the
effect of temporal frame count, our dual-head supervision, the $Z$-embedding component,
the residual fusion block, and the class-wise morphological closing augmentation.

Table~\ref{tab:temporal_ablation} isolates the effect of the number of BEV frames used
by the encoder. Increasing from one to three frames improves RayIoU from 29.5 to 30.8,
while FPS remains nearly unchanged (19.32 to 19.01). This indicates that LiNO does not
obtain its efficiency only by processing fewer frames, because the temporal aggregation is a
lightweight BEV-level operation and the occupancy head and runtime profile remain
stable across these settings.

\begin{table}[!t]
\centering
\small
\renewcommand{\arraystretch}{1.0}
\caption{Temporal frame-count variation.}
\label{tab:temporal_ablation}
\setlength{\tabcolsep}{5pt}
\begin{tabular}{l|ccc}
\hline
 & 1 frame & 2 frames & 3 frames \\
\hline
RayIoU $\uparrow$ & 29.5 & 30.3 & \textbf{30.8} \\
FPS $\uparrow$ & \textbf{19.32} & 19.16 & 19.01 \\
\hline
\end{tabular}
\end{table}

Tab.~\ref{tab:ablation_zembed_reduced} studies vertical embeddings. Without
$Z$-embedding, LiNO achieves a RayIoU of 27.1. Introducing 16 embedding channels
improves performance to 27.5 (+0.4), which is meaningful given the narrow margin
typically observed between competing methods in this benchmark. Using only 8 channels
results in a slight degradation (26.8), suggesting that insufficient vertical capacity
limits performance. Adding residual fusion on top of the 16-channel setup further
increases RayIoU from 27.5 to 29.0 (+1.5). This consistent gain indicates that residual
fusion enhances feature integration, leading to more coherent occupancy estimations.

\begin{table}[!h]
\centering
\small
\renewcommand{\arraystretch}{1.0}
\caption{$Z$-embedding and Residual fusion impacts to LiNO model.}
\label{tab:ablation_zembed_reduced}
\setlength{\tabcolsep}{5pt}
\begin{tabular}{c|c|c}
\hline
$Z$-emb. & Res-fus. & RayIoU $\uparrow$ \\
\hline
not & not & 27.1 \\
8 & not & 26.8 \\
16 & not & \textbf{27.5} \\
\hline
16 & not  & 27.5 \\
16 & yes & \textbf{29.0} \\
\hline
\end{tabular}
\end{table}

In Table~\ref{tab:ablation_closing_reduced}, using the (5,5,1) and (3,3,3) structures,
we observe that moderate closing ratios (0.2-0.5) improve performance, with a peak
RayIoU of 30.8 at a ratio of 0.2. We observed that mild closing mitigates
fragmentation in sparse regions and improves free-space continuity, leading to +0.9
RayIoU, while large ratios risk oversmoothing boundaries and slightly degrading
performance.

\begin{table}[!h]
\centering
\small
\renewcommand{\arraystretch}{1.0}
\caption{Morphological closing ratio.}
\label{tab:ablation_closing_reduced}
\setlength{\tabcolsep}{5pt}
\begin{tabular}{c|c}
\hline
Closing ratio & RayIoU $\uparrow$ \\
\hline
0.0 & 29.0  \\
0.2 & \textbf{30.8}   \\
0.5 & 29.9  \\
0.8 & 30.6  \\
1.0 & 30.4  \\
\hline
\end{tabular}
\end{table}

Table~\ref{tab:ablation_dualhead} compares three supervision configurations. Applying
the U-Net head directly on the FPN features together with the auxiliary head,
$\mathcal{H}_{U}(\mathbf{F}_{\text{FPN}})+\mathcal{H}_{aux}(\mathbf{F}_{\text{FPN}})$,
yields the best accuracy (31.32), indicating that hierarchical supervision on the same
feature manifold stabilizes optimization and improves voxel-level refinement. Placing
the auxiliary supervision inside the U-Net or duplicating it is less effective,
suggesting that the auxiliary signal is most beneficial when aligned with the features
processed by the final decoder.

\begin{table}[!h]
\centering
\small
\renewcommand{\arraystretch}{1.0}
\caption{Head configuration.}
\label{tab:ablation_dualhead}
\setlength{\tabcolsep}{5pt}
\begin{tabular}{c|c}
\hline
Dual-Head setup & RayIoU $\uparrow$ \\
\hline
$\mathcal{H}_{U}(\mathcal{H}_{aux}(\mathbf{F}_{\text{fpn}}))$ & 30.12 \\
$\mathcal{H}_{U}(\mathcal{H}_{aux}(\mathbf{F}_{\text{fpn}})) + \mathcal{H}_{aux}(\mathbf{F}_{\text{fpn}})$ & 29.56  \\
$\mathcal{H}_{U}(\mathbf{F}_{\text{fpn}}) + \mathcal{H}_{aux}(\mathbf{F}_{\text{fpn}})$ & \textbf{31.32} \\
\hline
\end{tabular}%
\end{table}

Finally, Table~\ref{tab:mask_aug} evaluates different kernel shapes for the data
augmentation. The second mask is fixed along the z-axis, since the first mask is applied
to volumetric objects and the second to flat objects, such as roads and sidewalks. We
tested different planar and cubic kernel shapes, considering that the ground truth has a
resolution of $200\times200\times16$ voxels. We keep the dimensions consistent to ensure
isotropic smoothing and avoid introducing directional bias into the occupancy structure.

\begin{table}[!h]
\centering
\small
\renewcommand{\arraystretch}{1.0}
\caption{Masks for augmentation.}
\label{tab:mask_aug}
\setlength{\tabcolsep}{5pt}
\begin{tabular}{c|c}
\hline
Structure shape & RayIoU $\uparrow$ \\
\hline
$(3\times3\times3)+(3\times3\times1)$  & 27.78 \\
$(5\times5\times5)+(5\times5\times1)$  & 27.82 \\
$(7\times7\times7)+(7\times7\times1)$  & 27.75 \\
$(3\times3\times3)+(5\times5\times1)$  & \textbf{28.40} \\
$(5\times5\times5)+(3\times3\times1)$  & 27.84 \\
\hline
\end{tabular}
\end{table}

\subsection{Efficiency Analysis}
\label{subsec:efficiency_analysis}

Table~\ref{tab:model_param_breakdown} summarizes the parameter distribution across the
main components of our model, while Table~\ref{tab:latency_mapping} reports the
corresponding runtime breakdown. Most parameters originate from the image backbone and its
feature extraction modules, followed by the \gls{lino} head and the radar-based point
backbone. Only the key elements of each subsystem are included, highlighting the modules
directly related to our contributions. For clarity, finer-grained submodule details are
omitted here, but these can be found in the original papers describing each respective
component \cite{huang2022bevdet4d, kim2023crncameraradarnet,
bispo2025panpillarsattentionbasednetwork3d, he2015deepresiduallearningimage,
zhu2020deformable, lang2019pointpillarsfastencodersobject}. Overall, our full model
contains 39.6 M trainable parameters, balancing expressiveness with computational
efficiency.

\begin{table}[!h]
\centering
\small
\renewcommand{\arraystretch}{1.0}
\caption{Parameter breakdown for LiNO-R50.}
\label{tab:model_param_breakdown}
\setlength{\tabcolsep}{3pt}
\begin{tabular}{l|l|r}
\hline
Name & Type & Params \\
\hline
backbone\_img\_module & RVTLSSFPN & 31.6M \\
- img\_backbone & ResNet & 23.5M \\
- img\_neck & SECONDFPN & 2.0M \\
- depth\_net & DepthNet & 5.4M \\
- view\_aggregation\_net & ViewAggregation & 807K \\
\hline
fuser & MFAFuser & 1.2M \\
\hline
backbone\_pts & PtsBackbone & 2.5M \\
- pts\_voxel\_encoder & PillarFeatureNet & 2.3K \\
- pillar\_att\_based & \gls{pan} & 1.1M \\
- pred\_cont\_occ & Sequential & 1.3M \\
\hline
head & LiNOHead & 4.3M \\
- bev\_backbone & CustomResNet3D & 3.0M \\
- bev\_neck & FPN3D & 27.8K \\
- z\_embed & ZEmbedding & 2.0K \\
- aux\_head\_conv & Conv3d & 1.7K \\
- bev\_head & U-Net & 1.2M \\
\hline
\multicolumn{2}{r|}{Trainable params} & 39.6M \\
\multicolumn{2}{r|}{Non-trainable params} & 9.5K \\
\multicolumn{2}{r|}{Total params} & 39.6M \\
\hline
\end{tabular}%
\end{table}

\begin{table}[!h]
\centering
\small
\renewcommand{\arraystretch}{1.0}
\caption{Latency mapping of the main modules considering LiNO-R50.}
\label{tab:latency_mapping}
\setlength{\tabcolsep}{6pt}
\begin{tabular}{l|r}
\hline
Module & Latency (ms) \\
\hline
img\_backbone & 7.74 \\
img\_dep & 1.42 \\
img\_transform & 9.31 \\
img\_pool & 0.48 \\
\textbf{total img} & 19.82 \\
\hline
pts\_voxelize & 2.04 \\
pts\_backbone & 3.29 \\
pts\_head & 0.74 \\
\textbf{total pts} & 6.59 \\
\hline
fusion\_pre & 0.80 \\
fusion\_layer & 5.03 \\
fusion\_post & 0.07 \\
\textbf{total fusion} & 6.33 \\
\hline
head\_backbone & 10.69 \\
head\_head & 14.62 \\
\textbf{total head} & 25.31 \\
\hline
\textbf{total} & 58.05 \\
\hline
\end{tabular}
\end{table}


Table~\ref{tab:latency_mapping} reports the latency breakdown of the main components of
our model using ResNet-50. The image branch accounts for 19.82~ms, dominated by the
image backbone and the geometric view transformation. The point-based branch (radar
branch) is substantially lighter, requiring only 6.59~ms in total. Fusion contributes an
additional 6.33~ms, with most of the cost concentrated in the fusion layer. The
occupancy head is the most expensive stage, totaling 25.31~ms due to the 3D backbone and
decoder operations. Overall, the full system achieves 58.05~ms per frame (17.2~FPS),
demonstrating that the proposed model remains computationally efficient across all
modules.

\subsection{Per-Class RayIoU Analysis}
\label{subsec:perclas}

\begin{figure}[!h]
  \centering
  \subfloat[Per-class RayIoU results on Occ3D using LiNO-R50. \label{fig:occ_rayiou_a}]{%
    \includegraphics[width=0.97\columnwidth]{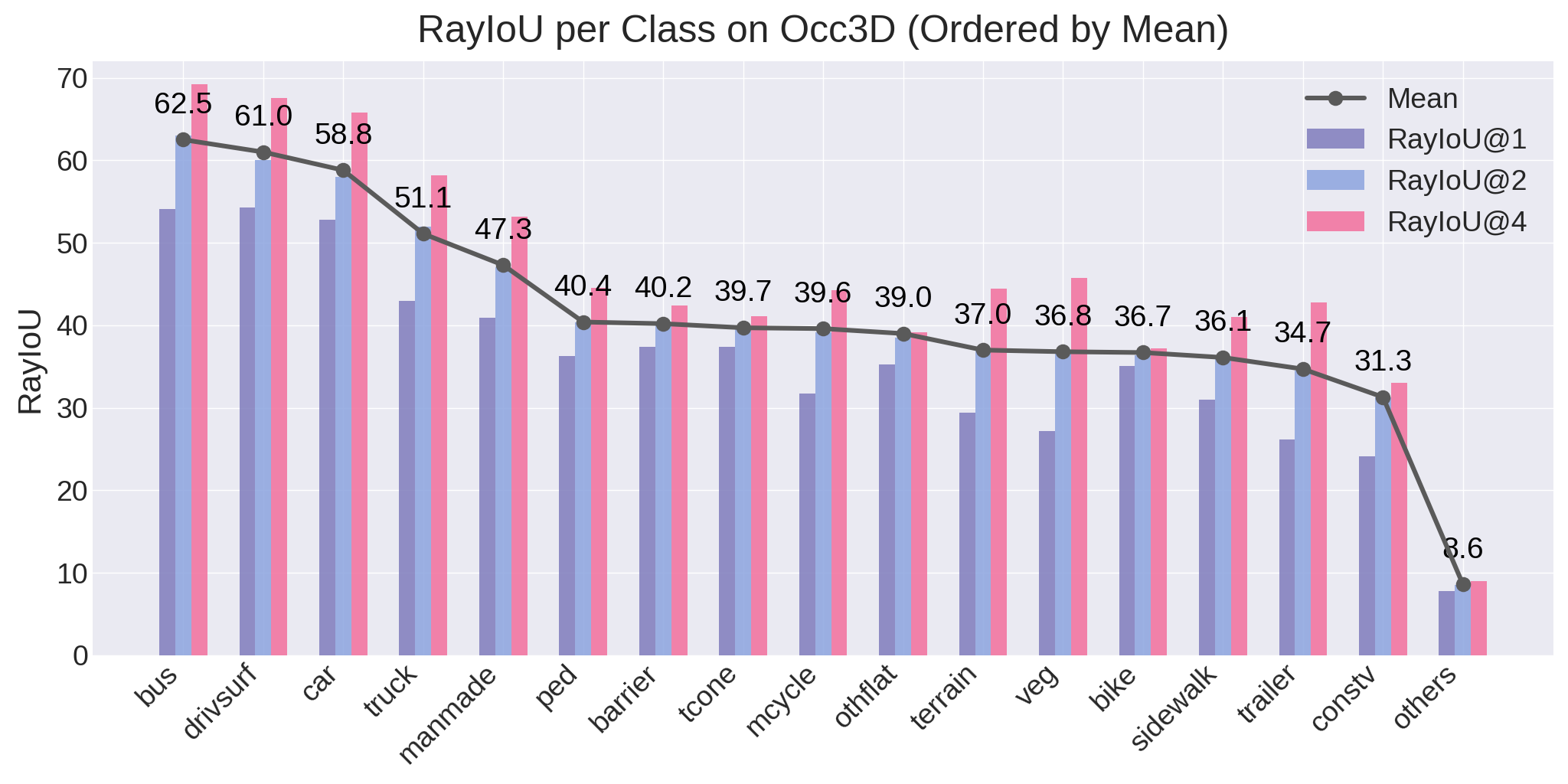}
  }
  \vspace{0.5em}
  \subfloat[Per-class RayIoU results on OpenOcc using LiNO-R50. \label{fig:occ_rayiou_b}]{%
    \includegraphics[width=0.97\columnwidth]{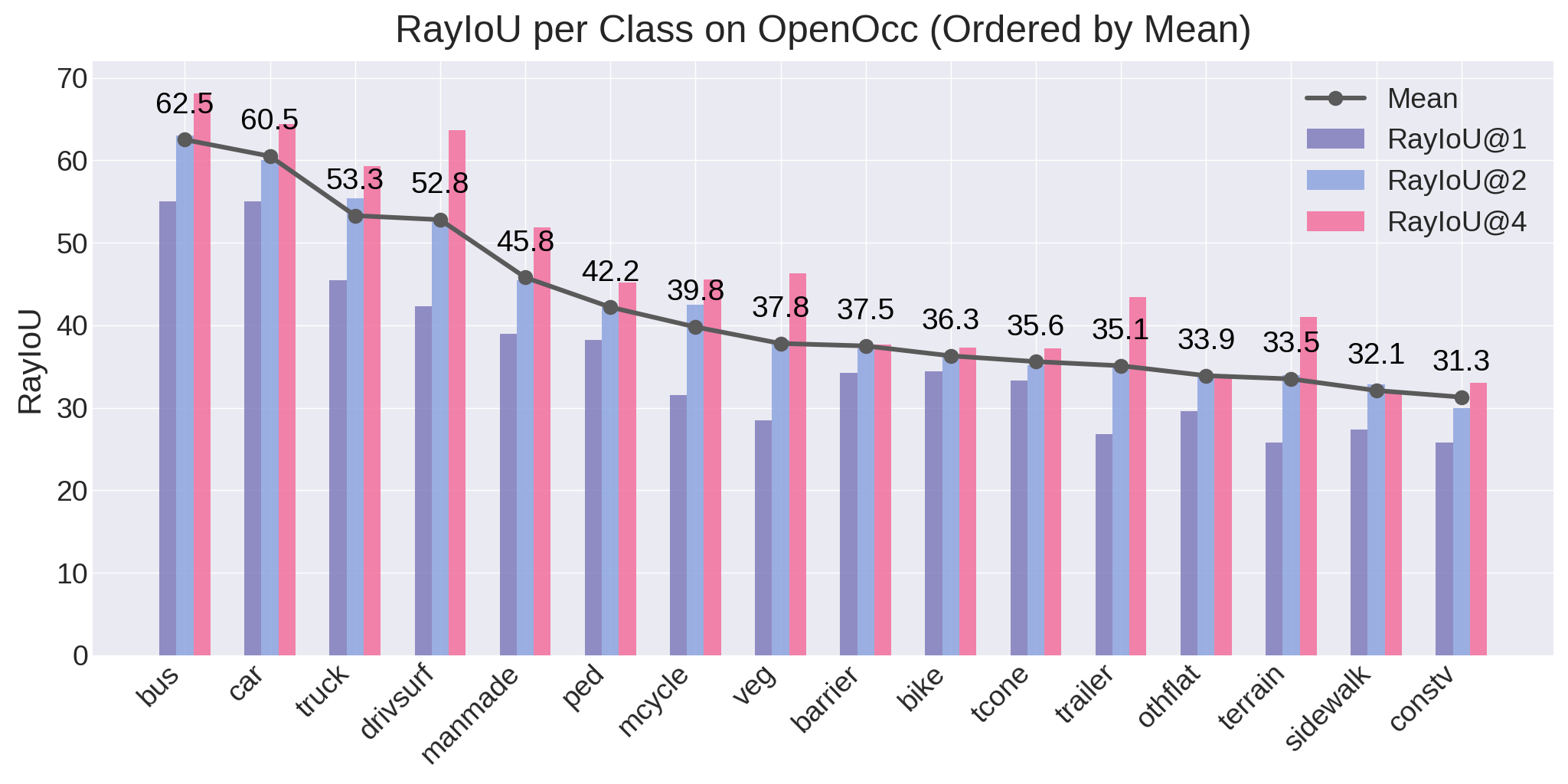}
  }
  \caption{
    Per-class RayIoU results for the Occ3D \cite{tian2023occ3d}
    (Fig.~\ref{fig:occ_rayiou_a}) and OpenOcc
    \cite{sima2023sceneoccupancy} (Fig.~\ref{fig:occ_rayiou_b}) datasets.
    Each class is evaluated under three distance tolerances: RayIoU@1, RayIoU@2,
    and RayIoU@4 (1~m, 2~m, and 4~m). Black numbers above each marker denote the
    mean RayIoU per class across the three thresholds. As the distance tolerance
    increases, the metric becomes more permissive, leading to higher RayIoU values
    as larger spatial errors are still counted as correct matches.
  }
  \label{fig:occ_rayiou}
\end{figure}

Figure~\ref{fig:occ_rayiou} presents the per-class \gls{rayiou} \cite{tang2024sparseocc}
results for both the Occ3D~\cite{tian2023occ3d} and
OpenOcc~\cite{sima2023sceneoccupancy} datasets. The black numbers above each marker
indicate the mean RayIoU for the corresponding class. Larger and more regularly shaped
classes, such as cars and buses, consistently achieve higher RayIoU scores. In contrast,
smaller or more irregular classes, such as traffic cones and pedestrians, tend to show
lower performance, reflecting the inherent difficulty of estimating fine-grained 3D
structures from sparse and noisy labels. These trends highlight the sensitivity of
RayIoU to class geometry and label sparsity, which affects thin or low-occupancy objects
more strongly.

\subsection{Qualitative Results}

\begin{figure*}[t!]
  \centering
  \includegraphics[width=1.0\linewidth]{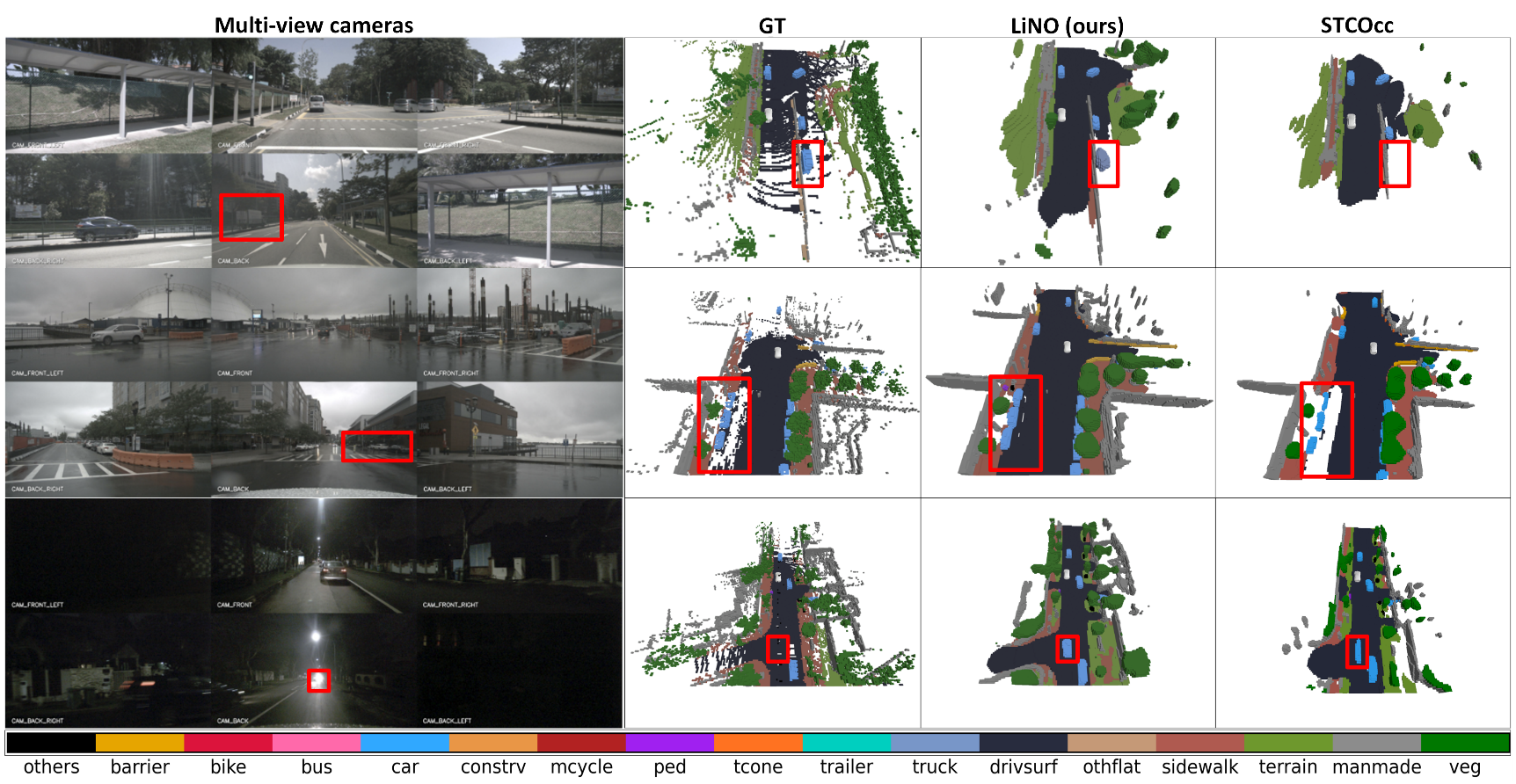}
  \caption{
  Comparison on the Occ3D \cite{tian2023occ3d}. LiNO accurately reconstructs occupancy maps even under GT errors (see camera views). Despite slightly lower RayIoU, LiNO produces visually denser and more consistent voxel predictions than the baseline, which may result from GT sparsity and the sensitivity of the RayIoU to mislabels (red boxes).
  }
  \label{fig:qual_occ}
\end{figure*}

As shown in Fig.~\ref{fig:qual_occ}, LiNO produces visually denser and more complete
occupancy reconstructions, accurately capturing voxels that are partially missing in the
labels. However, such plausible predictions can be penalised as false positives in
RayIoU, leading to lower numerical scores despite higher perceptual quality. This effect
can also be observed in the choice of dataset, as our model achieves SOTA results on
OpenOcc~\cite{sima2023sceneoccupancy}, which contains fewer empty voxels compared to
Occ3D~\cite{tian2023occ3d}. This suggests that, while RayIoU is a step forward over the
mIoU for evaluating volumetric consistency, it may still require refinement to better
account for label sparsity and visually coherent completions.

\section{Conclusions}
\label{sec:conclusions}
This work presented LiNO, a lightweight camera-radar architecture for 3D semantic
occupancy estimation. By combining compact multimodal BEV fusion with a
dual-supervised volumetric occupancy head, LiNO achieves strong reconstruction
accuracy while maintaining a deployment-oriented computational footprint. The
method reaches 17.2~FPS using 5.3~GB of GPU memory, compared with 4.1~FPS and
7.7~GB for STCOcc, while achieving state-of-the-art RayIoU on OpenOcc (41.9) and
mIoU on Occ3D (45.7).

These results demonstrate that accurate occupancy estimation does not necessarily
require a computationally expensive model. LiNO is explicitly designed around deployment constraints. Its main objective is to jointly address occupancy accuracy, inference latency,
and GPU memory consumption rather than optimizing only one of these criteria. This trade-off is particularly relevant for autonomous driving platforms, where occupancy estimation must operate alongside perception, planning, and control modules under limited computational and memory budgets.



\bibliography{ref}

@String(CVPR= {IEEE Conf. Comput. Vis. Pattern Recog.})

@String(ICCV= {Int. Conf. Comput. Vis.})

@String(ECCV= {Eur. Conf. Comput. Vis.})

@String(ICLR = {Int. Conf. Learn. Represent.})

@String(CVPR  = {CVPR})

@String(ICCV  = {ICCV})

@String(ECCV  = {ECCV})

@String(ICLR  = {ICLR})

@inproceedings{sima2023sceneoccupancy,
  author = {Tong, Wenwen and Sima, Chonghao and Wang, Tai and Chen, Li and Wu, Silei and Deng, Hanming and Gu, Yi and Lu, Lewei and Luo, Ping and Lin, Dahua and others},
  title = {Scene as Occupancy},
  booktitle = {Proceedings of the IEEE/CVF International Conference on Computer Vision (ICCV)},
  year = {2023},
  pages = {8406--8415},
  doi = {10.1109/ICCV51070.2023.00772},
  url = {https://doi.org/10.1109/ICCV51070.2023.00772}
}

@inproceedings{bateni2020co,
  author = {Bateni, Soroush and Wang, Zhendong and Zhu, Yuankun and Hu, Yang and Liu, Cong},
  title = {Co-Optimizing Performance and Memory Footprint Via Integrated CPU/GPU Memory Management, an Implementation on Autonomous Driving Platform},
  booktitle = {2020 IEEE Real-Time and Embedded Technology and Applications Symposium (RTAS)},
  year = {2020},
  pages = {310--323},
  doi = {10.1109/RTAS48715.2020.00007},
  url = {https://doi.org/10.1109/RTAS48715.2020.00007}
}

@article{wang2025multiple,
  title={Multiple attention mechanism for camera-radar fusion object detection},
  author={Wang, Wenbo and Wang, Yuwen and Zhang, Jun and Zhang, Weibin},
  journal={IEEE Transactions on Vehicular Technology},
  volume={74},
  number={5},
  pages={7416--7429},
  year={2025},
  publisher={IEEE},
  doi={10.1109/TVT.2025.3531379},
  url = {https://doi.org/10.1109/TVT.2025.3531379}
}

@misc{wang2024movingforwardreviewautonomous,
  author = {Wang, Xu and Maleki, Mohammad Ali and Azhar, Muhammad Waqar and Trancoso, Pedro},
  title = {Moving Forward: A Review of Autonomous Driving Software and Hardware Systems},
  year = {2024},
  eprint = {2411.10291},
  archivePrefix = {arXiv},
  primaryClass = {cs.RO},
  url = {https://arxiv.org/abs/2411.10291}
}

@techreport{singh2015critical,
  title={Critical reasons for crashes investigated in the national motor vehicle crash causation survey},
  author={Singh, Santokh},
  year={2015},
  address={Washington, DC, USA},
  url={https://crashstats.nhtsa.dot.gov/Api/Public/Publication/812506}
}

@article{yoo2018mems,
  title={MEMS-based lidar for autonomous driving},
  author={Yoo, Han Woong and Druml, Norbert and Brunner, David and Schwarzl, Christian and Thurner, Thomas and Hennecke, Marcus and Schitter, Georg},
  journal={Elektrotechnik und Informationstechnik},
  volume={135},
  number={6},
  pages={408--415},
  year={2018},
  doi={10.1007/s00502-018-0635-2},
  url={https://doi.org/10.1007/s00502-018-0635-2}
}

@article{li2020lidar,
  title={Lidar for Autonomous Driving: The Principles, Challenges, and Trends for Automotive Lidar and Perception Systems},
  author={Li, You and Ibanez-Guzman, Javier},
  journal={IEEE Signal Processing Magazine},
  volume={37},
  number={4},
  pages={50--61},
  year={2020},
  publisher={IEEE},
  doi={10.1109/MSP.2020.2973615},
  url={https://doi.org/10.1109/MSP.2020.2973615}
}

@article{MUELLER2020310,
title = {What humanlike errors do autonomous vehicles need to avoid to maximize safety?},
journal = {Journal of Safety Research},
volume = {75},
pages = {310-318},
year = {2020},
issn = {0022-4375},
doi = {https://doi.org/10.1016/j.jsr.2020.10.005},
url = {https://www.sciencedirect.com/science/article/pii/S0022437520301262},
author = {Alexandra S. Mueller and Jessica B. Cicchino and David S. Zuby},
}

@inproceedings{wolters2025unleashinghydrahybridfusion,
  author = {Wolters, Philipp and Gilg, Johannes and Teepe, Torben and Herzog, Fabian and Laouichi, Anouar and Hofmann, Martin and Rigoll, Gerhard},
  title = {Unleashing HyDRa: Hybrid Fusion, Depth Consistency and Radar for Unified 3D Perception},
  booktitle = {2025 IEEE International Conference on Robotics and Automation (ICRA)},
  year = {2025},
  doi = {10.1109/ICRA55743.2025.11127412},
  url = {https://doi.org/10.1109/ICRA55743.2025.11127412}
}

@inproceedings{cao2022monoscene,
  author = {Cao, Anh-Quan and de Charette, Raoul},
  title = {MonoScene: Monocular 3D Semantic Scene Completion},
  booktitle = {Proceedings of the IEEE/CVF Conference on Computer Vision and Pattern Recognition (CVPR)},
  year = {2022},
  pages = {3991--4001},
  doi = {10.1109/CVPR52688.2022.00396},
  url = {https://doi.org/10.1109/CVPR52688.2022.00396}
}

@inproceedings{huang2023tri,
  author = {Huang, Yuanhui and Zheng, Wenzhao and Zhang, Yunpeng and Zhou, Jie and Lu, Jiwen},
  title = {Tri-Perspective View for Vision-Based 3D Semantic Occupancy Prediction},
  booktitle = {Proceedings of the IEEE/CVF Conference on Computer Vision and Pattern Recognition (CVPR)},
  year = {2023},
  pages = {9223--9232},
  doi = {10.1109/CVPR52729.2023.00890},
  url = {https://doi.org/10.1109/CVPR52729.2023.00890}
}

@inproceedings{Zhang_2023_ICCV,
  author = {Zhang, Yunpeng and Zhu, Zheng and Du, Dalong},
  title = {OccFormer: Dual-Path Transformer for Vision-Based 3D Semantic Occupancy Prediction},
  booktitle = {Proceedings of the IEEE/CVF International Conference on Computer Vision (ICCV)},
  year = {2023},
  pages = {9433--9443},
  doi = {10.1109/ICCV51070.2023.00865},
  url = {https://doi.org/10.1109/ICCV51070.2023.00865}
}

@inproceedings{tang2024sparseocc,
  author = {Tang, Pin and Wang, Zhongdao and Wang, Guoqing and Zheng, Jilai and Ren, Xiangxuan and Feng, Bailan and Ma, Chao},
  title = {SparseOcc: Rethinking Sparse Latent Representation for Vision-Based Semantic Occupancy Prediction},
  booktitle = {Proceedings of the IEEE/CVF Conference on Computer Vision and Pattern Recognition (CVPR)},
  year = {2024},
  pages = {15035--15044},
  doi = {10.1109/CVPR52733.2024.01424},
  url = {https://doi.org/10.1109/CVPR52733.2024.01424}
}

@article{mei2024camera,
  author = {Mei, Jianbiao and Yang, Yu and Wang, Mengmeng and Zhu, Junyu and Ra, Jongwon and Ma, Yukai and Li, Laijian and Liu, Yong},
  title = {Camera-Based 3D Semantic Scene Completion With Sparse Guidance Network},
  journal = {IEEE Transactions on Image Processing},
  year = {2024},
  volume = {33},
  pages = {5468--5481},
  doi = {10.1109/TIP.2024.3461989},
  url = {https://doi.org/10.1109/TIP.2024.3461989}
}

@inproceedings{li2023voxformer,
  author = {Li, Yiming and Yu, Zhiding and Choy, Christopher and Xiao, Chaowei and Alvarez, Jose M. and Fidler, Sanja and Feng, Chen and Anandkumar, Anima},
  title = {VoxFormer: Sparse Voxel Transformer for Camera-Based 3D Semantic Scene Completion},
  booktitle = {Proceedings of the IEEE/CVF Conference on Computer Vision and Pattern Recognition (CVPR)},
  year = {2023},
  pages = {9087--9098},
  doi = {10.1109/CVPR52729.2023.00877},
  url = {https://doi.org/10.1109/CVPR52729.2023.00877}
}

@inproceedings{jiang2024symphonize,
  author = {Jiang, Haoyi and Cheng, Tianheng and Gao, Naiyu and Zhang, Haoyang and Lin, Tianwei and Liu, Wenyu and Wang, Xinggang},
  title = {Symphonize 3D Semantic Scene Completion With Contextual Instance Queries},
  booktitle = {Proceedings of the IEEE/CVF Conference on Computer Vision and Pattern Recognition (CVPR)},
  year = {2024},
  pages = {20258--20267},
  doi = {10.1109/CVPR52733.2024.01915},
  url = {https://doi.org/10.1109/CVPR52733.2024.01915}
}

@inproceedings{liao2025stcocc,
  author = {Liao, Zhimin and Wei, Ping and Chen, Shuaijia and Wang, Haoxuan and Ren, Ziyang},
  title = {STCOcc: Sparse Spatial-Temporal Cascade Renovation for 3D Occupancy and Scene Flow Prediction},
  booktitle = {Proceedings of the IEEE/CVF Conference on Computer Vision and Pattern Recognition (CVPR)},
  year = {2025},
  pages = {1516--1526},
  eprint = {2504.19749},
  archivePrefix = {arXiv},
  url = {https://arxiv.org/abs/2504.19749}
}

@article{li2024bevformer,
  author = {Li, Zhiqi and Wang, Wenhai and Li, Hongyang and Xie, Enze and Sima, Chonghao and Lu, Tong and Yu, Qiao and Dai, Jifeng},
  title = {BEVFormer: Learning Bird's-Eye-View Representation From LiDAR-Camera Via Spatiotemporal Transformers},
  journal = {IEEE Transactions on Pattern Analysis and Machine Intelligence},
  year = {2025},
  volume = {47},
  number = {3},
  pages = {2020--2036},
  doi = {10.1109/TPAMI.2024.3515454},
  url = {https://doi.org/10.1109/TPAMI.2024.3515454}
}

@misc{li2023fb,
  author = {Li, Zhiqi and Yu, Zhiding and Austin, David and Fang, Mingsheng and Lan, Shiyi and Kautz, Jan and Alvarez, Jose M.},
  title = {FB-OCC: 3D Occupancy Prediction Based on Forward-Backward View Transformation},
  year = {2023},
  note = {CVPR Workshop on End-to-End Autonomous Driving},
  eprint = {2307.01492},
  archivePrefix = {arXiv},
  url = {https://arxiv.org/abs/2307.01492}
}

@misc{huang2022bevdet4d,
  author = {Huang, Junjie and Huang, Guan},
  title = {BEVDet4D: Exploit Temporal Cues in Multi-Camera 3D Object Detection},
  year = {2022},
  eprint = {2203.17054},
  archivePrefix = {arXiv},
  primaryClass = {cs.CV},
  url = {https://arxiv.org/abs/2203.17054}
}

@inproceedings{ma2024cotr,
  author = {Ma, Qihang and Tan, Xin and Qu, Yanyun and Ma, Lizhuang and Zhang, Zhizhong and Xie, Yuan},
  title = {COTR: Compact Occupancy TRansformer for Vision-Based 3D Occupancy Prediction},
  booktitle = {Proceedings of the IEEE/CVF Conference on Computer Vision and Pattern Recognition (CVPR)},
  year = {2024},
  pages = {19936--19945},
  doi = {10.1109/CVPR52733.2024.01884},
  url = {https://doi.org/10.1109/CVPR52733.2024.01884}
}

@article{wang2024opus,
  author = {Wang, Jiabao and Liu, Zhaojiang and Meng, Qiang and Yan, Liujiang and Wang, Ke and Yang, Jie and Liu, Wei and Hou, Qibin and Cheng, Ming-Ming},
  title = {Opus: occupancy prediction using a sparse set},
  journal = {Advances in Neural Information Processing Systems},
  year = {2024},
  volume = {37},
  pages = {119861--119885},
  url = {https://proceedings.neurips.cc/paper_files/paper/2024/hash/d8ca28a32c05cd3b9b0940e43720f31b-Abstract-Conference.html}
}

@inproceedings{pan2024renderocc,
  author = {Pan, Mingjie and Liu, Jiaming and Zhang, Renrui and Huang, Peixiang and Li, Xiaoqi and Xie, Hongwei and Wang, Bing and Liu, Li and Zhang, Shanghang},
  title = {RenderOcc: Vision-Centric 3D Occupancy Prediction With 2D Rendering Supervision},
  booktitle = {2024 IEEE International Conference on Robotics and Automation (ICRA)},
  year = {2024},
  pages = {12404--12411},
  doi = {10.1109/ICRA57147.2024.10611537},
  url = {https://doi.org/10.1109/ICRA57147.2024.10611537}
}

@article{bispo2025panpillarsattentionbasednetwork3d,
  author = {Bispo, Ruan and Mitrev, Dane and Mariotti, Letizia and Botty, Clément and Humphrey, Denver and Scanlan, Anthony and Eising, Ciarán},
  title = {{PAN}: Pillars-Attention-Based Network for 3D Object Detection},
  journal = {IEEE Open Journal of Vehicular Technology},
  year = {2026},
  volume = {7},
  pages = {1--15},
  doi = {10.1109/OJVT.2026.3692626},
  url = {https://doi.org/10.1109/OJVT.2026.3692626}
}

@inproceedings{kim2023crncameraradarnet,
  author = {Kim, Youngseok and Shin, Juyeb and Kim, Sanmin and Lee, In-Jae and Choi, Jun Won and Kum, Dongsuk},
  title = {CRN: Camera Radar Net for Accurate, Robust, Efficient 3D Perception},
  booktitle = {Proceedings of the IEEE/CVF International Conference on Computer Vision (ICCV)},
  year = {2023},
  pages = {17569--17580},
  doi = {10.1109/ICCV51070.2023.01615},
  url = {https://doi.org/10.1109/ICCV51070.2023.01615}
}

@inproceedings{he2015deepresiduallearningimage,
  author = {He, Kaiming and Zhang, Xiangyu and Ren, Shaoqing and Sun, Jian},
  title = {Deep Residual Learning for Image Recognition},
  booktitle = {2016 IEEE Conference on Computer Vision and Pattern Recognition (CVPR)},
  year = {2016},
  pages = {770--778},
  doi = {10.1109/CVPR.2016.90},
  url = {https://doi.org/10.1109/CVPR.2016.90}
}

@inproceedings{lang2019pointpillarsfastencodersobject,
  author = {Lang, Alex H. and Vora, Sourabh and Caesar, Holger and Zhou, Lubing and Yang, Jiong and Beijbom, Oscar},
  title = {PointPillars: Fast Encoders for Object Detection From Point Clouds},
  booktitle = {2019 IEEE/CVF Conference on Computer Vision and Pattern Recognition (CVPR)},
  year = {2019},
  pages = {12689--12697},
  doi = {10.1109/CVPR.2019.01298},
  url = {https://doi.org/10.1109/CVPR.2019.01298}
}

@inproceedings{lin2017feature,
  author = {Lin, Tsung-Yi and Doll{\'a}r, Piotr and Girshick, Ross and He, Kaiming and Hariharan, Bharath and Belongie, Serge},
  title = {Feature Pyramid Networks for Object Detection},
  booktitle = {Proceedings of the IEEE Conference on Computer Vision and Pattern Recognition (CVPR)},
  year = {2017},
  pages = {936--944},
  doi = {10.1109/CVPR.2017.106},
  url = {https://doi.org/10.1109/CVPR.2017.106}
}

@inproceedings{philion2020lift,
  author = {Philion, Jonah and Fidler, Sanja},
  title = {Lift, Splat, Shoot: Encoding Images From Arbitrary Camera Rigs by Implicitly Unprojecting to 3D},
  booktitle = {Computer Vision -- ECCV 2020},
  year = {2020},
  pages = {194--210},
  doi = {10.1007/978-3-030-58568-6_12},
  url = {https://doi.org/10.1007/978-3-030-58568-6_12}
}

@article{lin2017focal,
  author = {Lin, Tsung-Yi and Goyal, Priya and Girshick, Ross and He, Kaiming and Doll{\'a}r, Piotr},
  title = {Focal Loss for Dense Object Detection},
  journal = {IEEE Transactions on Pattern Analysis and Machine Intelligence},
  year = {2020},
  volume = {42},
  number = {2},
  pages = {318--327},
  doi = {10.1109/TPAMI.2018.2858826},
  url = {https://doi.org/10.1109/TPAMI.2018.2858826}
}

@misc{lin2024teocc,
  author = {Lin, Zhiwei and Jin, Hongbo and Wang, Yongtao and Wei, Yufei and Dong, Nan},
  title = {TEOcc: Radar-Camera Multi-Modal Occupancy Prediction Via Temporal Enhancement},
  year = {2024},
  eprint = {2410.11228},
  archivePrefix = {arXiv},
  primaryClass = {cs.CV},
  url = {https://arxiv.org/abs/2410.11228}
}

@inproceedings{berman2018lovasz,
  author = {Berman, Maxim and Rannen Triki, Amal and Blaschko, Matthew B.},
  title = {The Lov{\'a}sz-Softmax Loss: A Tractable Surrogate for the Optimization of the Intersection-Over-Union Measure in Neural Networks},
  booktitle = {Proceedings of the IEEE Conference on Computer Vision and Pattern Recognition (CVPR)},
  year = {2018},
  pages = {4413--4421},
  doi = {10.1109/CVPR.2018.00464},
  url = {https://doi.org/10.1109/CVPR.2018.00464}
}

@inproceedings{li2019dice,
  author = {Li, Xiaoya and Sun, Xiaofei and Meng, Yuxian and Liang, Junjun and Wu, Fei and Li, Jiwei},
  title = {Dice Loss for Data-Imbalanced NLP Tasks},
  booktitle = {Proceedings of the 58th Annual Meeting of the Association for Computational Linguistics},
  year = {2020},
  pages = {465--476},
  doi = {10.18653/v1/2020.acl-main.45},
  url = {https://doi.org/10.18653/v1/2020.acl-main.45}
}

@article{zhang2018generalized,
  author = {Zhang, Zhilu and Sabuncu, Mert},
  title = {Generalized cross entropy loss for training deep neural networks with noisy labels},
  journal = {Advances in neural information processing systems},
  year = {2018},
  volume = {31},
  url = {https://proceedings.neurips.cc/paper_files/paper/2018/file/f2925f97bc13ad2852a7a551802feea0-Paper.pdf}
}

@inproceedings{lee2015deeply,
  author = {Lee, Chen-Yu and Xie, Saining and Gallagher, Patrick and Zhang, Zhengyou and Tu, Zhuowen},
  title = {Deeply-supervised nets},
  booktitle = {Artificial intelligence and statistics},
  year = {2015},
  pages = {562--570},
  organization = {Pmlr},
  url = {https://proceedings.mlr.press/v38/lee15a.html}
}

@inproceedings{szegedy2015going,
  author = {Szegedy, Christian and Liu, Wei and Jia, Yangqing and Sermanet, Pierre and Reed, Scott and Anguelov, Dragomir and Erhan, Dumitru and Vanhoucke, Vincent and Rabinovich, Andrew},
  title = {Going deeper with convolutions},
  booktitle = {Proceedings of the IEEE conference on computer vision and pattern recognition},
  year = {2015},
  pages = {1--9},
  url = {https://arxiv.org/abs/1409.4842}
}

@inproceedings{xie2015holistically,
  author = {Xie, Saining and Tu, Zhuowen},
  title = {Holistically-nested edge detection},
  booktitle = {Proceedings of the IEEE international conference on computer vision},
  year = {2015},
  pages = {1395--1403},
  url = {https://openaccess.thecvf.com/content_iccv_2015/html/Xie_Holistically-Nested_Edge_Detection_ICCV_2015_paper.html}
}

@inproceedings{tian2023occ3d,
 author = {Tian, Xiaoyu and Jiang, Tao and Yun, Longfei and Mao, Yucheng and Yang, Huitong and Wang, Yue and Wang, Yilun and Zhao, Hang},
 booktitle = {Advances in Neural Information Processing Systems},
 editor = {A. Oh and T. Naumann and A. Globerson and K. Saenko and M. Hardt and S. Levine},
 pages = {64318--64330},
 publisher = {Curran Associates, Inc.},
 title = {Occ3D: A Large-Scale 3D Occupancy Prediction Benchmark for Autonomous Driving},
 url = {https://proceedings.neurips.cc/paper_files/paper/2023/file/cabfaeecaae7d6540ee797a66f0130b0-Paper-Datasets_and_Benchmarks.pdf},
 volume = {36},
 year = {2023}
}

@inproceedings{caesar2020nuscenes,
  author = {Caesar, Holger and Bankiti, Varun and Lang, Alex H. and Vora, Sourabh and Liong, Venice Erin and Xu, Qiang and Krishnan, Anush and Pan, Yu and Baldan, Giancarlo and Beijbom, Oscar},
  title = {nuScenes: A Multimodal Dataset for Autonomous Driving},
  booktitle = {Proceedings of the IEEE/CVF Conference on Computer Vision and Pattern Recognition (CVPR)},
  year = {2020},
  pages = {11618--11628},
  doi = {10.1109/CVPR42600.2020.01164},
  url = {https://doi.org/10.1109/CVPR42600.2020.01164}
}

@misc{kingma2017adammethodstochasticoptimization,
  author = {Kingma, Diederik P. and Ba, Jimmy},
  title = {Adam: A Method for Stochastic Optimization},
  year = {2014},
  eprint = {1412.6980},
  archivePrefix = {arXiv},
  primaryClass = {cs.LG},
  url = {https://arxiv.org/abs/1412.6980}
}

@inproceedings{paszke2019pytorch,
 author = {Paszke, Adam and Gross, Sam and Massa, Francisco and Lerer, Adam and Bradbury, James and Chanan, Gregory and Killeen, Trevor and Lin, Zeming and Gimelshein, Natalia and Antiga, Luca and Desmaison, Alban and Kopf, Andreas and Yang, Edward and DeVito, Zachary and Raison, Martin and Tejani, Alykhan and Chilamkurthy, Sasank and Steiner, Benoit and Fang, Lu and Bai, Junjie and Chintala, Soumith},
 booktitle = {Advances in Neural Information Processing Systems},
 editor = {H. Wallach and H. Larochelle and A. Beygelzimer and F. d\textquotesingle Alch\'{e}-Buc and E. Fox and R. Garnett},
 pages = {},
 publisher = {Curran Associates, Inc.},
 title = {PyTorch: An Imperative Style, High-Performance Deep Learning Library},
 url = {https://proceedings.neurips.cc/paper_files/paper/2019/file/bdbca288fee7f92f2bfa9f7012727740-Paper.pdf},
 volume = {32},
 year = {2019}
}

@misc{zhang2023occnerf,
  author = {Zhang, Chubin and Yan, Juncheng and Wei, Yi and Li, Jiaxin and Liu, Li and Tang, Yansong and Duan, Yueqi and Lu, Jiwen},
  title = {OccNeRF: Self-Supervised Multi-Camera Occupancy Prediction With Neural Radiance Fields},
  year = {2023},
  eprint = {2312.09243},
  archivePrefix = {arXiv},
  url = {https://arxiv.org/abs/2312.09243}
}

@misc{yu2023flashocc,
  author = {Yu, Zichen and Shu, Changyong and Deng, Jiajun and Lu, Kangjie and Liu, Zongdai and Yu, Jiangyong and Yang, Dawei and Li, Hui and Chen, Yan},
  title = {FlashOcc: Fast and Memory-Efficient Occupancy Prediction Via Channel-to-Height Plugin},
  year = {2023},
  eprint = {2311.12058},
  archivePrefix = {arXiv},
  url = {https://arxiv.org/abs/2311.12058}
}

@inproceedings{shi2024occupancy,
  author = {Shi, Yiang and Cheng, Tianheng and Zhang, Qian and Liu, Wenyu and Wang, Xinggang},
  title = {Occupancy as set of points},
  booktitle = {European Conference on Computer Vision},
  year = {2024},
  pages = {72--87},
  organization = {Springer},
  url = {https://eccv.ecva.net/virtual/2024/poster/2470}
}

@inproceedings{li2024viewformer,
  author = {Li, Jinke and He, Xiao and Zhou, Chonghua and Cheng, Xiaoqiang and Wen, Yang and Zhang, Dan},
  title = {ViewFormer: Exploring Spatiotemporal Modeling for Multi-View 3D Occupancy Perception Via View-Guided Transformers},
  booktitle = {Computer Vision -- ECCV 2024},
  year = {2024},
  pages = {90--106},
  doi = {10.1007/978-3-031-72775-7_6},
  url = {https://doi.org/10.1007/978-3-031-72775-7_6}
}

@article{Xu_2025,
  author = {Xu, Huaiyuan and Chen, Junliang and Meng, Shiyu and Wang, Yi and Chau, Lap-Pui},
  title = {A survey on occupancy perception for autonomous driving: The information fusion perspective},
  journal = {Information Fusion},
  year = {2025},
  volume = {114},
  pages = {102671},
  publisher = {Elsevier BV},
  doi = {10.1016/j.inffus.2024.102671},
  url = {https://doi.org/10.1016/j.inffus.2024.102671},
  issn = {1566-2535},
  month = feb
}

@inproceedings{zhu2020deformable,
  author = {Zhu, Xizhou and Su, Weijie and Lu, Lewei and Li, Bin and Wang, Xiaogang and Dai, Jifeng},
  title = {Deformable DETR: Deformable Transformers for End-to-End Object Detection},
  booktitle = {International Conference on Learning Representations (ICLR)},
  year = {2021},
  url = {https://arxiv.org/abs/2010.04159}
}

@inproceedings{ronneberger2015u,
  author = {Ronneberger, Olaf and Fischer, Philipp and Brox, Thomas},
  title = {U-Net: Convolutional Networks for Biomedical Image Segmentation},
  booktitle = {Medical Image Computing and Computer-Assisted Intervention -- MICCAI 2015},
  year = {2015},
  pages = {234--241},
  doi = {10.1007/978-3-319-24574-4_28},
  url = {https://doi.org/10.1007/978-3-319-24574-4_28}
}

@inproceedings{hsiao2022zhuyi,
  author = {Hsiao, Yu-Shun and Hari, Siva Kumar Sastry and Filipiuk, Micha{\l} and Tsai, Timothy and Sullivan, Michael B. and Reddi, Vijay Janapa and Singh, Vasu and Keckler, Stephen W.},
  title = {Zhuyi: Perception Processing Rate Estimation for Safety in Autonomous Vehicles},
  booktitle = {Proceedings of the 59th ACM/IEEE Design Automation Conference},
  year = {2022},
  pages = {289--294},
  doi = {10.1145/3489517.3530445},
  url = {https://doi.org/10.1145/3489517.3530445}
}

@INPROCEEDINGS{zhang2020optimal,
  author={Zhang, Yajia and Sun, Hongyi and Zhou, Jinyun and Pan, Jiacheng and Hu, Jiangtao and Miao, Jinghao},
  booktitle={2020 IEEE Intelligent Vehicles Symposium (IV)}, 
  title={Optimal Vehicle Path Planning Using Quadratic Optimization for Baidu Apollo Open Platform}, 
  year={2020},
  volume={},
  number={},
  pages={978-984},
  doi = {10.1109/IV47402.2020.9304787},
  url = {https://doi.org/10.1109/IV47402.2020.9304787}
}

@ARTICLE{kilian2022emergency,
  author={Kilian, Philipp and Koller, Oliver and Van Bergen, Patrick and Gebauer, Carsten and Heidinger, Frederic and Dazer, Martin},
  journal={IEEE Access}, 
  title={Emergency Operation in the Power Supply Domain According to ISO 26262}, 
  year={2022},
  volume={10},
  number={},
  pages={47557-47569},
  doi = {10.1109/ACCESS.2022.3170903},
  url = {https://doi.org/10.1109/ACCESS.2022.3170903}
}

@inproceedings{xu2022aoi,
  author = {Xu, Chengyuan and Xu, Qian and Wang, Jianping and Wu, Kui and Lu, Kejie and Qiao, Chunming},
  title = {AoI-Centric Task Scheduling for Autonomous Driving Systems},
  booktitle = {IEEE INFOCOM 2022 -- IEEE Conference on Computer Communications},
  year = {2022},
  pages = {1019--1028},
  doi = {10.1109/INFOCOM48880.2022.9796958},
  url = {https://doi.org/10.1109/INFOCOM48880.2022.9796958}
}
\begin{IEEEbiography}[{\includegraphics[width=1in,height=1.3in,clip,keepaspectratio]{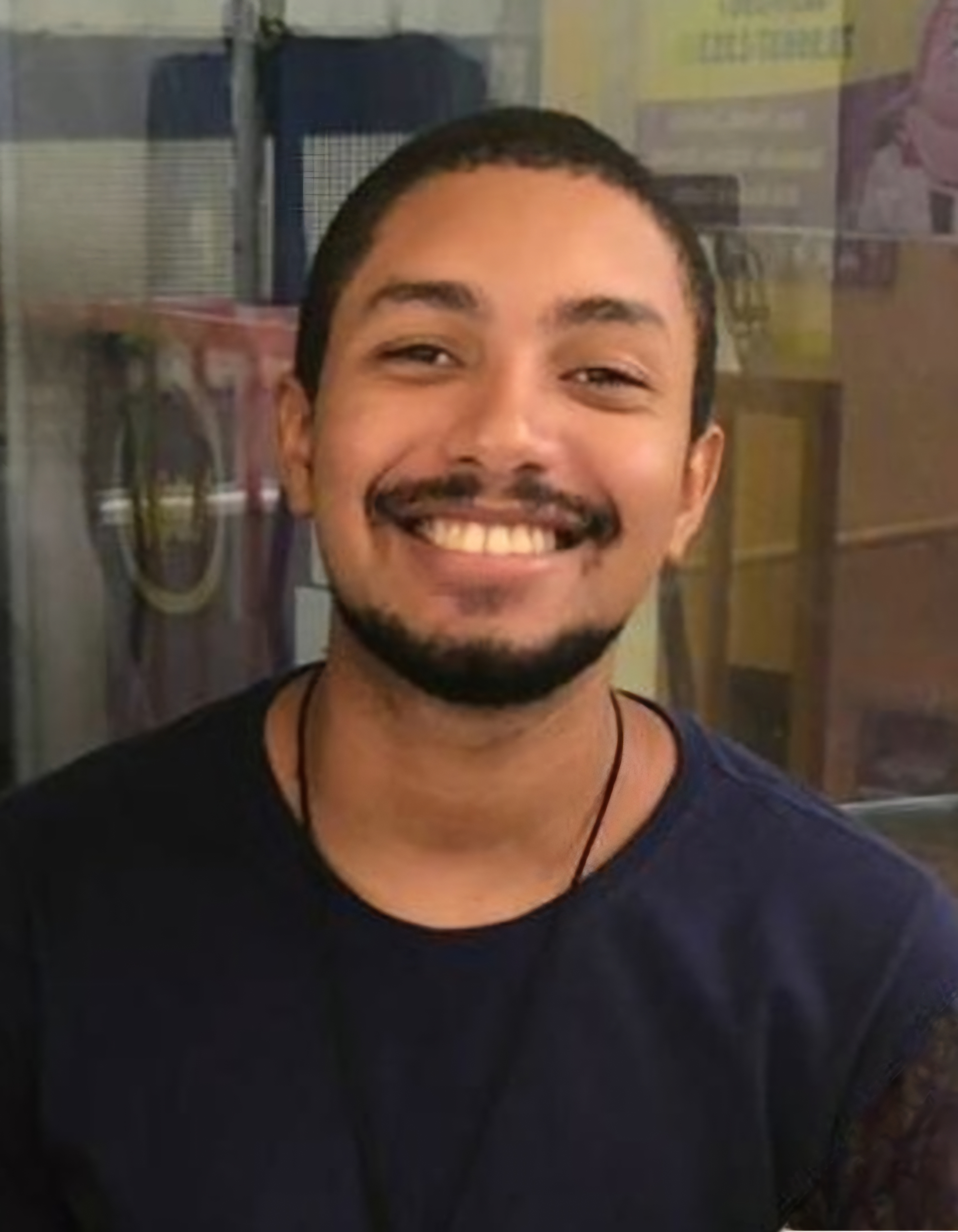}}]{Ruan Bispo }
received the BA degree in electronic engineering from the Federal University of Sergipe in 2020, the MSE degree in dynamic systems from the University of S\~ao Paulo in 2023, and is currently a second-year PhD student in computer vision and artificial intelligence at the University of Limerick. He was a researcher at Ford Motor Company (2 years) and a data scientist at Samsung (2 years). His research interests include robotics, autonomous vehicles, control, and AI.
\end{IEEEbiography}

\begin{IEEEbiography}[{\includegraphics[width=1in,height=1.3in,clip,keepaspectratio]{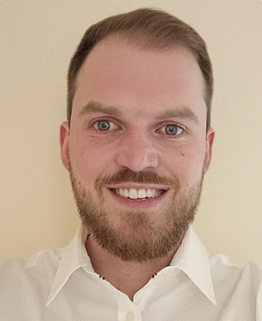}}]{Tim Brophy }
received the B.Eng. (Hons.) degree from the University of Galway in 2018. From 2018 to 2025, he was a member of the Connaught Automotive Research (CAR) Group at the University of Galway, where he completed his PhD studies. In 2025, he joined the University of Limerick as a Postdoctoral Researcher. His research interests include sensor availability, computer vision, and artificial intelligence in the context of automated vehicles.
\end{IEEEbiography}

\begin{IEEEbiography}[{\includegraphics[width=1in,height=1.3in,clip,keepaspectratio]{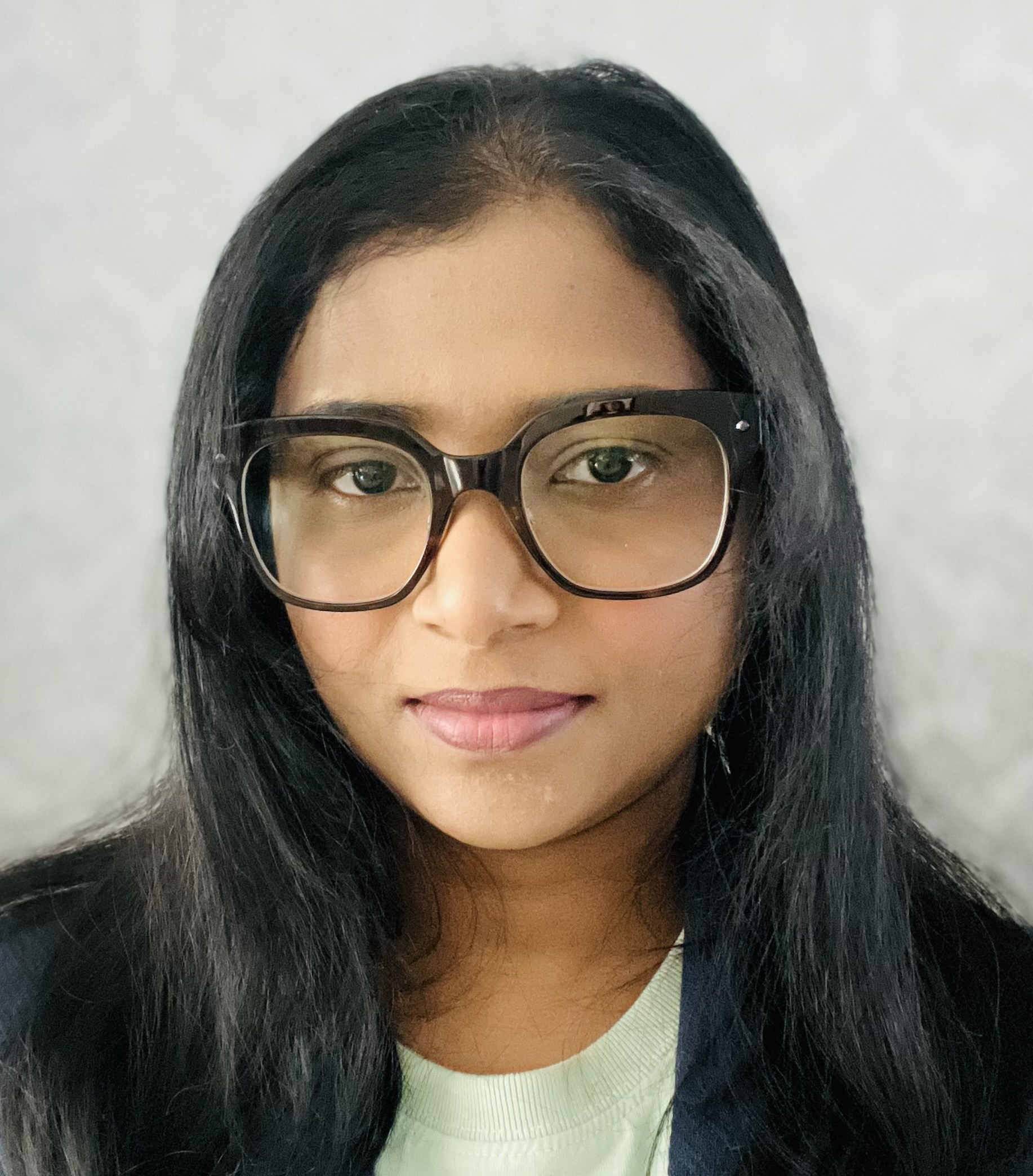}}]{Reenu Mohandas }
received the M.Tech. degree in digital image computing from the University of Kerala, India, in 2014, the M.Sc. degree in Artificial Intelligence from CIT, currently Munster Technological University, Cork, Ireland, in 2019, and Ph.D. from the Department of Electronic and Computer Engineering (ECE), University of Limerick, Ireland, and currently a Postdoctoral Researcher there. Her research interests include computer vision, deep learning, incremental learning, VQA models, and quantisation and edge deployment of deep learning algorithms.
\end{IEEEbiography}

\begin{IEEEbiography}[{\includegraphics[width=1in,height=1.3in,clip,keepaspectratio]{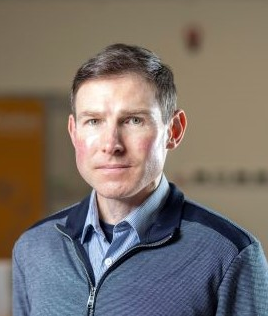}}]{\textbf{Anthony Scanlan }} received the B.Sc. degree in experimental physics from the National University of Ireland Galway, Galway, Ireland, in 1998 and the M.Eng. and Ph.D. degrees in electronic engineering from the University of Limerick, Limerick, Ireland, in 2001 and 2005, respectively. He is currently a Senior Research Fellow at the Dept. of Electronic \& Computer Engineering, University of Limerick, Ireland, and has been the principal investigator on several research projects in the areas of signal processing and data converter design. His current research interests are in the areas of artificial intelligence, computer vision, and industrial and environmental applications.
\end{IEEEbiography}

\begin{IEEEbiography}[{\includegraphics[width=1in,height=1.3in,clip,keepaspectratio]{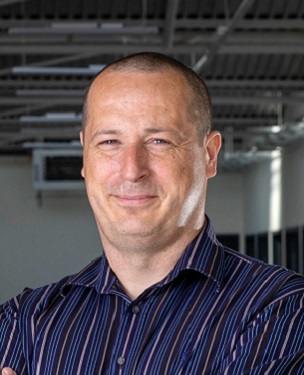}}]{Ciarán Eising } received the BE in Electronic and Computer Engineering and a PhD from the NUI Galway, in 2003 and 2010, respectively. From 2009 to 2020, he was a Computer Vision Architect \& Senior Expert with Valeo. In 2020, he joined the University of Limerick as an Associate Professor.
\end{IEEEbiography}

\end{document}